\documentclass[final]{cvpr}

\usepackage{times}
\usepackage{epsfig}
\usepackage{graphicx}
\usepackage{amsmath}
\usepackage{amssymb}

\usepackage[dvipsnames]{xcolor}
\usepackage{subfigure}
\usepackage{enumerate}
\usepackage{symbols} 
\usepackage{multirow}
\usepackage{enumerate}
\usepackage{soul}
\usepackage[title]{appendix}

\newcommand{\X}{\mathcal{X}}
\newcommand{\Y}{\mathcal{Y}}
\newcommand{\y}{\Y}
\newcommand{\x}{\X}
\newcommand{\dy}{\delta \Y}

\newcommand{\D}{\mathcal{D}}
\newcommand{\E}{\mathbb{E}}

\newcommand{\lin}{\text{lin}}

\newtheorem{prop}{Proposition}

\usepackage[pagebackref=true,breaklinks=true,colorlinks,bookmarks=false]{hyperref}
\usepackage{cleveref}

\def\cvprPaperID{10041}
\def\confYear{CVPR 2021}

\begin{document}

\title{A linearized framework and a new benchmark for model selection for fine-tuning}

\author{Aditya Deshpande, Alessandro Achille, Avinash Ravichandran, Hao Li, Luca Zancato, \\
Charless Fowlkes, Rahul Bhotika, Stefano Soatto and Pietro Perona\\
Amazon Web Services\\
{\tt\small \{deshpnde,aachille,ravinash,haolimax,zancato,fowlkec,bhotikar,soattos,peronapp\}@amazon.com}
}

\maketitle

\begin{abstract}
Fine-tuning from a collection of models pre-trained on different domains (a ``model zoo'') is emerging 
as a technique to improve test accuracy in the low-data regime. However, model 
selection, \ie how to pre-select the right model to fine-tune from a model zoo without performing 
any training, remains an open topic. We use a linearized framework to approximate fine-tuning, 
and introduce two new baselines for model selection -- Label-Gradient and 
Label-Feature Correlation. Since all model selection algorithms in the literature have been 
tested on different use-cases and never compared directly, we introduce a new comprehensive 
benchmark for model selection comprising of: $i)$ A model zoo of single and multi-domain models, and $ii)$ 
Many target tasks. Our benchmark highlights accuracy gain with model zoo compared to
fine-tuning Imagenet models. We show our model selection baseline can 
select optimal models to fine-tune in few selections and has the highest ranking correlation to 
fine-tuning accuracy compared to existing algorithms. 
\end{abstract}


\section{Introduction}

\begin{figure*}
    \vspace{-.5cm}
    \centering
    \subfigure[{\bf Model zoo vs.\ different architectures.} Fine-tuning using our \textcolor{OliveGreen}{model zoo} is better (\ie lower test error) than fine-tuning using different architectures with \textcolor{red}{Random} or \textcolor{blue}{Imagenet pre-trained} initialization. We use fine-tuning hyper-parameters of \secref{sec:finetune}
    with $\eta=.005$.] {\includegraphics[width=.46\textwidth]{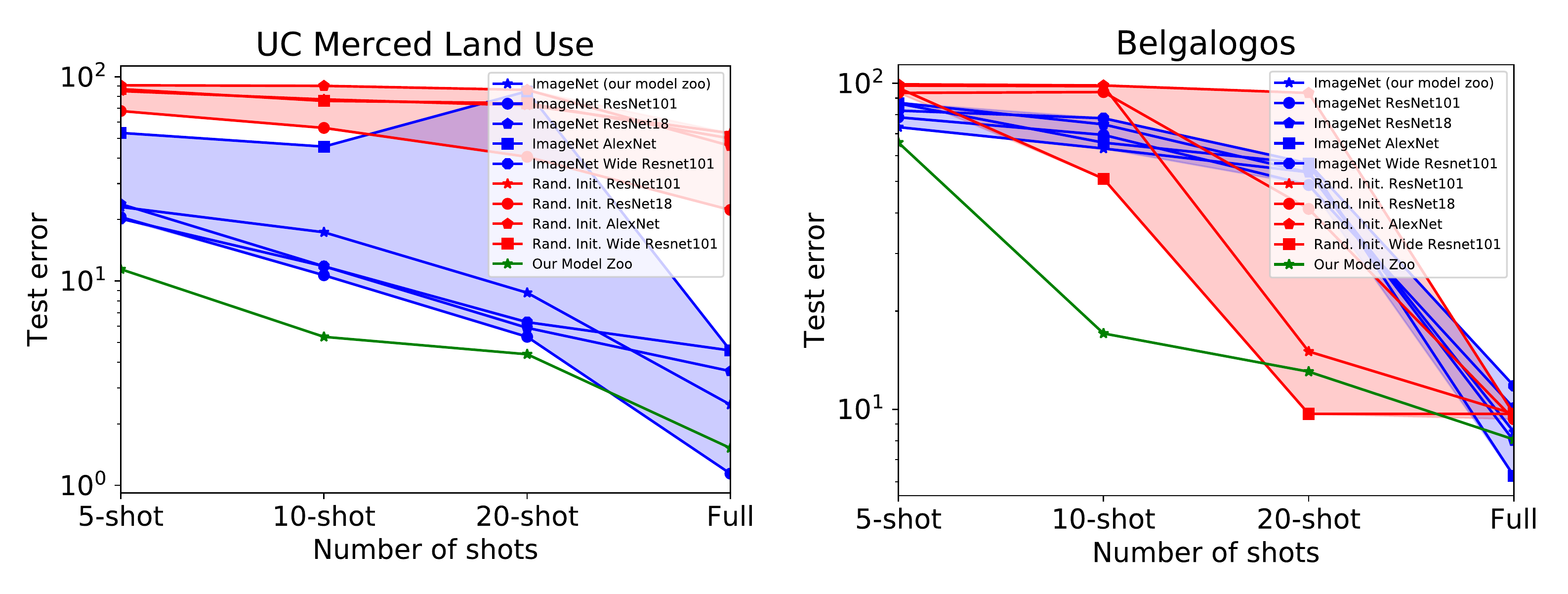}} \hspace{.2cm}
    \subfigure[{\bf Model zoo vs.\ HPO. of Imagenet expert} Fine-tuning using our \textcolor{OliveGreen}{model zoo} is better than fine-tuning with hyper-parameter optimization (HPO) on \textcolor{blue}{Imagenet pre-trained} Resnet-101 model. We use fine-tuning hyper-parameters of~\secref{sec:finetune} and perform HPO with $\eta = .01, .005, 0.001$.]{\includegraphics[width=.5\textwidth]{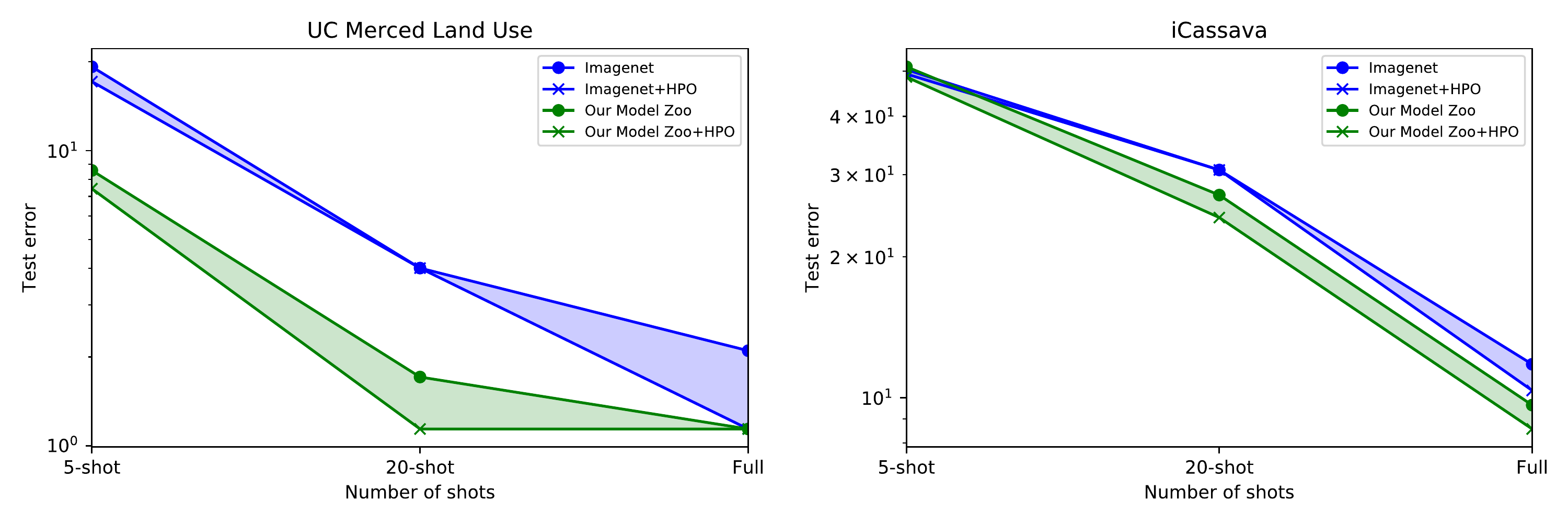}
    }
    \caption{
    \textbf{Fine-tuning using our model zoo can obtain lower test error compared to: $(a)$ using different architectures and $(b)$ hyper-parameter optimization (HPO) of Imagenet expert.}
    The standard fine-tuning approach entails picking a network architecture pre-trained on Imagenet to fine-tune and performing hyper-parameter optimization (HPO) during fine-tuning. We outperform this strategy by fine-tuning using our model zoo described in \secref{sec:model_zoo}. We plot test error as a function of 
    the number of per-class samples (\ie shots) in the dataset. In $(a)$, we compare fine-tuning with our single-domain experts in the model zoo to using different 
    architectures (AlexNet, ResNet-18, ResNet-101, Wide ResNet-101) for fine-tuning.
    In $(b)$, we show fine-tuning with our model zoo obtains lower error than performing HPO on Imagenet pre-trained Resnet-101~\cite{7780459} during fine-tuning. 
    Model zoo lowers the test error, especially in the low-data regime (5, 10, 20-shot per class samples of target task). 
    Since we compare to Imagenet fine-tuning, we exclude Imagenet experts from our model zoo for the above plots. 
    }
    \vspace{-.5cm}
    \label{fig:efficiency-curve}
\end{figure*}

A ``{\bf model zoo}'' is a collection of pre-trained models, obtained by training different architectures 
on many datasets covering a variety of tasks and domains. For instance, the zoo could comprise models (or experts) trained to 
classify, say, trees, birds, fashion items, aerial images, etc. The typical use of a model zoo is to provide a 
good initialization which can then be fine-tuned for a new target task, for which we have few training data. This 
strategy is an alternative to the more common practice of starting from a model trained on a large dataset, 
say Imagenet \cite{imagenet_cvpr09}, and is aimed at providing better domain coverage and a stronger inductive 
bias. Despite the growing usage of model zoos~\cite{cui2018large,big_transfer,li2020rethinking,48798} 
there is little in the way of analysis, both theoretical and empirical, to illuminate which approach 
is preferable under what conditions. In \figref{fig:efficiency-curve}, we show that fine-tuning with 
a model zoo is indeed better, especially when training data is limited. \figref{fig:efficiency-curve}
also shows that using a model zoo, we can outperform hyper-parameter optimization performed during 
fine-tuning of the Imagenet pre-trained model.


Fine-tuning with a model zoo can be done by brute-force fine-tuning of each model in the zoo, or more 
efficiently by using ``{\bf model selection}'' to select the closest model (or best initialization) 
from which to fine-tune. The goal of model selection therefore is to find the best pre-trained model
to fine-tune on the target task, without performing the actual fine-tuning. 
So, we seek an approximation to the fine-tuning process. In our work, we develop an analytical 
framework to characterize the fine-tuning process using a linearization of the model around 
the point of pre-training \cite{mu2020gradients}, drawing inspiration from the work on the Neural 
Tangent Kernel (NTK) \cite{jacot2018neural,lee2019wide}. Our analysis of generalization bounds and 
training speed using linearized fine-tuning naturally suggests two criterion to select the best model 
to fine-tune from, which we call Label-Gradient Correlation (LGC) and Label-Feature Correlation (LFC). 
Given its simplicity, we consider our criteria as {\em baselines}, rather than full-fledged methods 
for model selection, and compare the state-of-the-art in model selection -- \eg RSA~\cite{DwivediR19}, 
LEEP~\cite{nguyen2020leep}, Domain Similarity~\cite{cui2018large}, Feature Metrics~\cite{ueno2020a} 
-- against it. 

Model selection being a relatively recent endeavor, there is currently no standard dataset 
or a common benchmark to perform such a comparison. For example, LEEP~\cite{nguyen2020leep} 
performs its model selection experiments on transfer (or fine-tuning) from Imagenet pre-trained 
model to $200$ randomly sampled tasks of CIFAR-100~\cite{Krizhevsky09learningmultiple} 
image classification, RSA~\cite{DwivediR19} uses the Taskonomy dataset~\cite{taskonomy2018} to evaluate 
its prediction of task transfer (or model selection) performance. Due to these different experimental setups, 
the state-of-the-art in model selection is unclear. Therefore, in \secref{sec:experiments} we
build a new benchmark comprising a large model zoo and many target tasks. For our model zoo, we 
use $8$ large image classification datasets (from different domains) to train single-domain 
and multi-domain experts. We use various image classification datasets as target tasks and 
study fine-tuning (\secref{sec:finetune}) and model selection (\secref{sec:model_selection}) using 
our model zoo. To the best of our knowledge ours is the first large-scale benchmark for model 
selection.


By performing fine-tuning and model selection on our benchmark, we discover the following:
\begin{enumerate}[(a)]
  \setlength{\itemsep}{0pt}
  \setlength{\parskip}{0pt}
  \item  We show (\figref{fig:efficiency-curve}) that fine-tuning models in the model zoo 
  can outperform the standard method of fine-tuning with Imagenet pre-trained architectures and 
  HPO. We obtain better fine-tuning than Imagenet expert with, both model zoo of 
  single-domain experts (\figref{fig:finetune_full}) and multi-domain experts (\figref{fig:finetune_universal}). 
  While in the high-data 
  regime using a model zoo leads to modest gains, it sensibly improves accuracy in the low-data regime. 
  \item For any given target task, we show that only a small subset of the models in the zoo 
  lead to accuracy gain (\figref{fig:finetune_full}). In such a scenario, brute-force fine-tuning all models to find the few 
  that improve accuracy is wasteful. Fine-tuning with all our single-domain experts in the 
  model zoo is $40\times$ more compute intensive than fine-tuning an Imagenet Resnet-101 expert 
  in \tabref{tab:runtime}. 
  \item Our LGC model selection, and particularly its approximation LFC, 
  can find the best models from which to fine-tune without requiring an expensive brute-force search
  (\tabref{tab:runtime}). With only $3$ selections, we can select models that show gain over Imagenet expert  
  (\figref{fig:model_zoo_selection_top3}). Compared to Domain Similarity~\cite{Cui2018iNatTransfer}, 
  RSA~\cite{DwivediR19} and Feature Metrics~\cite{ueno2020a}, our LFC score can select the best model 
  to fine-tune in fewer selections, and it shows the highest ranking correlation to the fine-tuning test accuracy 
  (\figref{fig:trials}) among all model selection methods.
  
\end{enumerate}


\section{Related work}
\textbf{Fine-tuning.} 
The exact role of pre-training and fine-tuning in deep learning is still debated. He et 
al.~\cite{he2019rethinking} show that, for \textit{object detection}, the 
accuracy of a pre-trained model can be matched by simply training a network from scratch but 
for longer. However, they notice that the pre-trained model is more robust to different 
hyper-parameters and outperforms training from scratch in the low-data regime. 
On the other hand, in \textit{fine-grained visual classification}, Li et al.~\cite{li2020rethinking}  show 
that even after hyper-parameter optimization (HPO) and with longer training, models pre-trained on 
similar tasks can significantly outperform both Imagenet pre-training and training from scratch. 
Achille et al.~\cite{achille2019task2vec}, Cui et al.~\cite{Cui2018iNatTransfer} study task 
similarity and also report improvement in performance by using the right pre-training. Zoph et 
al.~\cite{zoph2020rethinking} show that while pre-training is useful in low-data regime, 
self-training outperforms pre-training in high-data regime. Most of the above work, 
~\cite{achille2019information,Cui2018iNatTransfer,li2020rethinking} draws inferences
of transfer learning by using Imagenet~\cite{imagenet_cvpr09} or iNaturalist~\cite{HornASSAPB17} 
experts. We build a model zoo with many more single domain and multi-domain 
experts (\secref{sec:model_zoo}), and use various target tasks (\secref{sec:finetune}) to 
empirically study transfer learning in different data regimes.

\textbf{Model Selection.}
Empirical evidence \cite{achille2019task2vec,li2020rethinking,zamir2018taskonomy} and theory \cite{achille2019information} 
suggests that effectiveness of fine-tuning relates to a notion of distance between tasks. Taskonomy 
\cite{zamir2018taskonomy} defines a distance between learning tasks \textit{a-posteriori}, that is, by looking 
at the fine-tuning accuracy during transfer learning. However, for predicting the best pre-training without
performing fine-tuning, an \textit{a-priori} approach is best. Achille et al.~\cite{achille2019task2vec,achille2019information} 
introduce a fixed-dimensional ``task embedding'' to encode distance between tasks. Cui et al.~\cite{Cui2018iNatTransfer} 
propose a Domain Similarity measure, which entails using the Earth Mover Distance (EMD) between source and target 
features. LEEP~\cite{nguyen2020leep,tran2019transferability} looks at the conditional cross-entropy between the 
output of the pre-trained model and the target labels. RSA~\cite{DwivediR19} compares representation dissimilarity
matrices of features from pre-trained model and a small network trained on target task for model selection. As
opposed to using the ad-hoc measure of task similarity, we rely on a linearization approximation 
to the fine-tuning to derive our model selection methods (\secref{sec:approach}).

\textbf{Linearization and NTK.} To analyse fine-tuning from pre-trained weights, we use a simple but effective framework 
inspired by the Neural Tangent Kernel (NTK)  formalism \cite{jacot2018neural}: We approximate the fine-tuning dynamics by 
looking at a linearization of the source model around the pre-trained weights $w_0$ (\secref{sec:linearization}). 
This approximation has been suggested by \cite{mu2020gradients}, who also notes that while there 
may be doubts on whether an NTK-like approximation holds for real randomly-initialized network \cite{goldblum2019truth}, 
it is more likely to hold in the case of fine-tuning, since the fine-tuned weights tend to remain close to the pre-trained 
weights. 

\textbf{Few-shot.} Interestingly, while pre-training has a higher impact in the few-shot regime, there 
is only a handful of papers that experiment with it \cite{dvornik2020selecting,Goyal2019,triantafillou2019meta}. 
This could be due to over-fitting of the current literature on standard benchmarks that have a restricted scope. 
We hope that our proposed benchmark (\secref{sec:experiments}) may foster further research.



\section{Approach}
\label{sec:approach}

{\bf Notation.} We have a model zoo, $\mathcal{F}$, of 
$n$ pre-trained models or experts: $\mathcal{F} = \{f^{1}, f^{2}, \cdots f^{n}\}$. 
Our aim is to classify a target dataset, $\mathcal{D} = \{(x_i,y_i)\}_{i=1}^{N}$,
by fine-tuning models in the model zoo. Here, $x_i \in \cX$, is the $i^{th}$ input image and 
$y_i \in \cY$, is the corresponding class label. 
For a network $f \in \mathcal{F}$ with weights $w$, we denote the output of the network with  $f_{w}(x)$. $w_0$ denotes the initialization (or pre-trained weights) of
models in the model zoo. The goal of model selection is to predict
a score $S(f_{w_{0}}, \mathcal{D})$ that measures the fine-tuning accuracy 
on the test set $\mathcal{D}^{\text{test}}$, when $\mathcal{D}$ is used to fine-tune 
the model $f_{w_0}$. Note, $S$ does not have to exactly measure 
the fine-tuning accuracy, it needs to only predict a score that correlates to the
ranking by fine-tuning accuracy. The model selection score for every pre-trained model, 
$S(f^{k}, \mathcal{D})$ for $k \in \{1,2,\cdots,n\}$, 
can then be used as proxy to rank and select top-$k$ models by their 
fine-tuning accuracy. Since the score $S$ needs to estimate (a proxy for) fine-tuning
accuracy without performing any fine-tuning, in \secref{sec:linearization}
we construct a linearization approximation to fine-tuning and present several
results that allow us to derive our \emph{Label-Gradient Correlation} ($S_{LG}$) and 
\emph{Label-Feature Correlation} ($S_{LF}$) (\secref{sec:label-correlation}) 
scores for model selection from it. In~\figref{fig:trials} (b), we show
our scores have higher ranking correlation to fine-tuning accuracy than existing work.
 
\subsection{Linearized framework to analyse fine-tuning}
\label{sec:linearization}


Given an initialization $w_0$, the weights of the pre-trained model, we can define the linearized model:
\[
f^\lin_w(x) := f_{w_0}(x) +\nabla_w f_{w_0}(x)|_{w=w_0} (w - w_0),
\]
which approximates the output of the real model for $w$ close to $w_0$. Mu et al.~\cite{mu2020gradients} 
observe that, while in general not accurate, a linear approximation can correctly describe 
the model throughout fine-tuning since the weights $w$ tend to remain close to the 
initial value $w_0$. Under this linear approximation \cite{lee2019wide} shows the following proposition,

\begin{prop}
\vspace{-.2cm}
\label{prop:ntk-dynamics}
Let $\D = \{(x_i, y_i)\}_{i=1}^N$ be the target dataset. Assume the task is a binary classification problem with labels $y_i=\pm 1$,%
\footnote{This is to simplify the notation, but a similar result would hold for a multi-class classification using one-hot encoding. Using the $L_2$ loss is necessary to have a close form expression. However, note that empirically the $L_2$ performs similarly to cross-entropy during fine-tuning \cite{golik2013cross,barz2020deep}.}
using the $L_2$ loss $L_\D(w) = \sum_{i=1}^N (y_i - f_w(x_i))^2$. Let $w_t$ denote the weights at time $t$ during training. Then, the loss function evolve as:
\begin{align}
L_t &= (\Y - f_{w_0}(\X))^T e^{-2\eta \Theta t} (\Y - f_{w_0}(\X)) \label{eq:ntk-loss}
\end{align}
where $f_{w_0}(\X)$ denotes the vector containing the output of the network on all the images in the dataset, $\Y$ denotes the vectors of all training labels, and we defined the Neural Tangent Kernel (NTK) matrix:
\begin{equation}
\label{eq:ntk-definition}
\Theta := \nabla_w f_w(\X) \nabla_w f_w(\X)^T
\end{equation}
which is the $N\times N$ Gram matrix of all the per-sample gradients.
\vspace{-.2cm}
\end{prop}

From Prop.~\ref{prop:ntk-dynamics}, the behavior of the network during fine-tuning is fully characterized by 
the kernel matrix $\Theta$, which  depends on the pre-trained model $f_{w_0}$,  the data $\mathcal{X}$ and 
the task labels $\y$. We then expect to be able to select the best model by looking at these quantities. 
To show how we can do this, we now derive several results connecting $\Theta$ and $\y$ to the quantities of relevance 
for model selection below, \ie Training time and Generalization on the target task.

\textbf{Training time.}
In ~\cite{zancato2020predicting}, it is shown that the loss $L_{t}$ of the 
linearized model evolves with training over time $t$ as
\begin{equation}
L_t = \|\dy\|^2 - t \dy^T \Theta \dy + O(t^2).
\label{eq:convergence-time-first-order}
\end{equation}
where we have defined $\dy = \y - f_{w_0}(\x)$ to be the initial projection residual. Eq.~(\ref{eq:convergence-time-first-order})
suggests using the quadratic term $\dy^T \Theta \dy$ as a simple estimate of the training speed.

\textbf{Generalization.}
The most important criterion for model selection is generalization performance. Unfortunately, we cannot have any close form characterization of generalization error, which depends on test data we do not have. However, in \cite{arora2019fine} the following bound on the test error is suggested:

\begin{equation}
L_\text{test}^2 \leq \frac{1}{n} \y^T \Theta^{-1} \y = \frac{1}{n} \sum_{k} \frac{1}{\lambda_k} (\y \cdot v_k)^2.
\label{eq:ntk-generalization}
\end{equation}

We see that if $\y$ correlates more with the first principal components of variability of the per-sample gradients (so that $\y \cdot v_k$ is larger), then we expect better generalization.

Arora et al.~\cite{arora2019fine} prove that this bound holds with high-probability for a wide-enough randomly initialized 3-layer network. In practice,
however, this generalization bound may be vacuous as hypotheses are not satisfied (the network is deeper, and the initialization is not Gaussian).  For this reason, rather than using the above quantity as a real bound, we refer to it as an empirical ``generalization score''.

Note \cref{eq:convergence-time-first-order} and \cref{eq:ntk-generalization} contain the similar terms $\dy^T \Theta \dy$ and $\dy^T \Theta^{-1} \dy$. By diagonalizing $\Theta$ and applying Jensen's inequality we have the following relation between the two:

\begin{equation}
\Big(\frac{\dy^T \Theta \dy}{\|\dy\|^2} \Big)^{-1} \leq  \frac{\dy^T \Theta^{-1} \dy}{\|\dy\|^2}.
\label{eq:generalization-speed-bound}
\end{equation}

Hence, good ``generalization score'' $\dy^T \Theta^{-1} \dy$ implies faster initial fine-tuning, that is, larger $\dy^T \Theta \dy$. In general 
we expect the two quantities to be correlated. Hence, selecting the fastest model to train or the one that generalizes better are correlated 
objectives. $\y^T \Theta \y$ is an approximation to $\dy^T \Theta \dy$ that uses task labels $\y$ and kernel $\Theta$, and we use
it to derive our model selection scores in \secref{sec:label-correlation}. Large value of $\y^T \Theta \y$ implies
better generalization and faster training and it is desirable for a model when fine-tuning. 

{\bf Should model selection use gradients or features?} Our analysis is in terms of 
the matrix $\Theta$ which depends on the network’s gradients \eqref{eq:ntk-definition}, 
not on its features. In~\secref{sec:proof}, we show that it suffices to use 
features (\ie network activations) in \eqref{eq:ntk-definition} as an approximation 
to the NTK matrix. Let $\lbrack f(x_i)\rbrack_{l}$ denote the feature vector (or activation) 
extracted from layer $l$ of pre-trained network $f$ after forward pass on image, 
\ie after $f(x_i)$. In analogy with the gradient similarity matrix $\Theta$ of 
\eqref{eq:ntk-definition}, we define the feature similarity matrix $\Theta_F$ (which approximates $\Theta$) as 
follows
\begin{equation}
\label{eq:ntk-feature-definition}
\Theta_F := \lbrack f_w(\X) \rbrack_{l} \lbrack f_w(\X) \rbrack_{l}^T.
\end{equation}

\subsection{Label-Feature and Label-Gradient correlation}
\label{sec:label-correlation}

We now introduce our two scores for model selection, \emph{Label-Gradient correlation} and 
\emph{Label-Feature correlation}.

\textbf{Label-Gradient Correlation.} From \secref{sec:linearization} we know that the following 
score, 
\begin{equation}
S_\text{LG}(f_{w_0}, \mathcal{D}) = \y^T \Theta \y = \Theta \cdot \y \y^T
\label{eq:label-gradient-dot-product}    
\end{equation}
which we call \emph{Label-Gradient Correlation} (LGC), can be used to estimate both the convergence 
time (eq.~\ref{eq:convergence-time-first-order}) and the generalization ability of a model. Here, 
``$\cdot$'' denotes the dot-product of the matrices (\ie the sum of Hadamard product of two matrices).
$\y \y^T$ is an $N\times N$ matrix such that $(\y \y^T)_{i,j} = 1$ if $x_i$ and $x_j$ have 
the same label and $-1$ otherwise. For this reason, we call $\y\y^T$ the label similarity 
matrix. On the other hand, $\Theta_{ij} = \nabla_w f_{w_0}(x_i) \cdot \nabla_w f{w_0}(x_j)$ is 
the pair-wise similarity matrix of the gradients. Hence, \cref{eq:label-gradient-dot-product} can be 
interpreted as giving high LG score (i.e., the model is good for the task) if the gradients are 
similar whenever the labels are also similar, and are different otherwise.



\textbf{Label-Feature Correlation.} Instead of $\Theta$, we can use the approximation 
$\Theta_F$ from \eqref{eq:ntk-feature-definition} and define our 
\emph{Label-Feature Correlation} (LFC) score as:
\[
S_\text{LF} = \y^T \Theta_F \y = \Theta_F \cdot \y\y^T.
\]
Similarly to the LGC score, this score is higher if samples with the same labels have similar features
extracted from the pre-trained network.

\subsection{Implementation} 
\label{sec:implementation} 

Notice that the scores $S_\text{LG}$ and $S_\text{LF}$ are not normalized. Different 
pre-training could lead to very different scores if the gradients or the features have a 
different norm. Also, $\y\y^T$ used in our scores is specific to binary classification. 
In practice, we address this as follows: For a multi-class classification problem, let 
$K_\y$ be the $N \times N$-matrix with $(K_{\y})_{i,j} =  1$ if $x_i$ and $x_j$ have 
the  same label, and $-1$ otherwise. Let $\mu_K$ denote the mean of the entries of 
$K_\y$, and $\mu_\Theta$ the mean of $\Theta$. We define the normalized LGC score as:

\begin{equation}
S_\text{LG} = \frac{(\Theta - \mu_\Theta) \cdot (K_\y - \mu_K) }{\|\Theta - \mu_\Theta\|_2 \|K_\y  - \mu_K\|_2},
\label{eq:lgc-pearson}
\end{equation}


We normalize LFC similar to LGC in \eqref{eq:lgc-pearson}. This can also be interpreted as 
the Pearson's Correlation coefficient between the entries of $\Theta$ (or $\Theta_F$) and the entries 
of $K_\y$, justifying the name label-gradient (or label-feature) correlation.

{\bf Which features and gradients to use?} For LFC, we extract features from the layer before 
the fully-connected classification layer (for both Resnet-101~\cite{7780459} and 
Densenet-169~\cite{HuangLW16a} models in our model zoo of \secref{sec:model_zoo}). 
We use these features to construct our $\Theta_F$ and compute the normalized LFC. 
For LGC, following~\cite{mu2020gradients}, we use gradients corresponding to the 
last convolutional layer in the pre-trained network. For a large gradient vector,
to perform fast computation of LGC, we take a random projection to $10K$ dimensions 
and compute the normalized LGC score. This results in a trade-off between accuracy 
and computation for LGC.

{\bf Sampling of target task.} Model selection is supposed to be an inexpensive
pre-processing step before actual fine-tuning. To reduce its computation, following 
previous work of RSA~\cite{DwivediR19}, we sample the training set of 
target dataset $\cD$ and pick at most $25$ images per class to compute our model selection scores. 
Note, test set is hidden from model selection. Our
results show, this still allows us to select models that obtain accuracy gain 
over Imagenet expert (\figref{fig:model_zoo_selection_top3}), and 
we need few selections ($<7$ for model zoo size $30$) to select the optimal 
models (\figref{fig:trials}) to fine-tune. We include additional implementation
details of our model selection methods and other baselines: RSA~\cite{DwivediR19},
Domain Similarity~\cite{Cui2018iNatTransfer}, LEEP~\cite{nguyen2020leep},
Feature Metrics~\cite{ueno2020a} in~\secref{sec:model_selection_impl}.

\begin{table*}[!t]
\vspace{-.6cm}
\begin{center}
\resizebox{\textwidth}{!}{
\setlength{\tabcolsep}{2pt}
\begin{tabular}{|l||c||cccccccc|}
\hline
{} & {\bf Pre-train} &  {\bf RESISC-45}~\cite{7891544} &  {\bf Food-101}~\cite{bossard14} &  {\bf Logo 2k}~\cite{643999} &  {\bf  G. Landmark}~\cite{8237636} &  {\bf iNaturalist 2019}~\cite{HornASSAPB17} &  {\bf iMaterialist}~\cite{imat} &  
{\bf ImageNet}~\cite{imagenet_cvpr09} &  {\bf Places-365}~\cite{zhou2017places} \\
\hline
\hline
\multirow{2}{*}{{\bf Densenet-169}}
& $\times$ &          93.61 &     82.38 &    64.58 &               82.28 &        71.34 &         66.59 &     76.40 &   55.47 \\
& $\checkmark$ &          96.34 &     87.82 &    76.78 &               84.89 &        73.65 &         67.57 &       - &   55.58 \\
\hline
\hline
\multirow{2}{*}{{\bf Resnet-101}}                         
& $\times$ &          87.14 &     79.20 &    62.03 &               78.48 &        70.32 &         67.95 &     {\bf 77.54} &   55.83 \\
& $\checkmark$ &          {\bf 96.53} &     {\bf 87.95} &    {\bf 78.52} &               {\bf 85.64} &        74.37 &         68.58 &       - &   {\bf 56.08} \\
\hline
\hline
{\bf Reported Acc.}     
& -     &          86.02~\cite{acc_nwpu_resisc45} &     86.99~\cite{acc_food-101} &    67.65~\cite{acc_logo_2k} &                 - &        {\bf 75.40}~\cite{acc_inaturalist} &           -
&     77.37\cite{acc_imagenet} &   54.74~\cite{zhou2017places} \\
\hline
\end{tabular}}
\end{center}
\caption{{\bf Model zoo of single-domain experts.} We train $30$ 
models, Resnet-101 and Densenet-169, on $8$ source datasets and measure the top-1 
test accuracy. We train our models starting with ($\checkmark$) and without 
($\times$) Imagenet pre-training. For all datasets we have higher test accuracy 
with Resnet-101 ($\checkmark$) than what is reported in the literature (last row), 
except for iNaturalist~\cite{HornASSAPB17} by -1.03\%. We order datasets from 
left to right by increasing dataset size, Nwpu-resisc45~\cite{7891544} 
has $25K$ training images while Places-365~\cite{zhou2017places} has $1.8M$.
We chose datasets that are publicly available and cover different domains.}
\label{tab:source_dataset}
\vspace{-.6cm}

\end{table*}

\begin{table}[!t]
\vspace{-.1cm}
\begin{center}
\resizebox{.5\textwidth}{!}{
\begin{tabular}{|l||c||ccc|}
\hline
{\bf Dataset} & {\bf Single Domain} & {\bf Shared} & {\bf Multi-BN}& {\bf Adapter} \\
\hline
\hline
Nwpu-resisc45~\cite{7891544} & 96.53 & 73.73 & 96.46 & 95.24 \\
Food-101~\cite{bossard14} & 87.95 & 48.12 & 87.92 & 86.35 \\
Logo 2k~\cite{643999} & 78.52 & 24.39 & 79.06 & 70.13 \\
Goog. Land~\cite{8237636} & 85.64 & 65.1 & 81.89 & 76.83 \\
iNatural.~\cite{HornASSAPB17} & 74.37 & 37.6 & 65.2 & 63.04 \\
iMaterial.~\cite{imat} & 68.58 & 42.15 & 63.27 & 57.5 \\
Imagenet~\cite{imagenet_cvpr09} & 77.54 & 52.51 & 69.03 & 58.9 \\
Places-365~\cite{zhou2017places} & 56.08 & 41.58 & 51.21 & 47.51 \\
\hline
\end{tabular}}
\end{center}
\caption{{\bf Multi-domain expert.} The top-1 test accuracy of multi-domain 
model -- Multi-BN, Adapter -- is comparable to single domain expert for small datasets (Nwpu-Resisc45, Food-101, 
Logo 2k), while the accuracy is lower on other large datasets. Multi-BN performs better than 
Shared, Adapter on all datasets and we use this as our multi-domain expert for fine-tuning
and model selection.}
\label{tab:multi_domain}
\vspace{-.6cm}
\end{table}

\section{Experiments}
\label{sec:experiments}

Having established the problem of model selection for fine-tuning (\secref{sec:approach}), 
we now put our techniques to test. \secref{sec:model_zoo} describes our construction
of model zoos with single-domain and multi-domain experts. In \secref{sec:finetune}, we then verify the advantage of 
fine-tuning using our model zoo with various target tasks. In~\secref{sec:model_selection}, 
we compare our LFC, LGC model selection (\secref{sec:label-correlation}) to previous 
work, and show that our method can select the optimal models to fine-tune from our model 
zoo (without performing the actual fine-tuning).

\subsection{Model Zoo}
\label{sec:model_zoo}

We evaluate model selection and fine-tuning with both, a model zoo of single-domain
experts (\ie models trained on single dataset) and a model zoo of multi-domain 
experts described below.

\textbf{Source Datasets.} \tabref{tab:source_dataset} and \tabref{tab:suppl_datasets} lists the source datasets,
\ie the datasets used for training our model zoo. We include publicly available large source 
datasets (from $25K$ to $1.8M$ training images) from different domains, 
\eg Nwpu-resisc45 \cite{7891544} consists of aerial imagery, Food-101~\cite{bossard14} 
and iNaturalist 2019~\cite{HornASSAPB17} consist of food, plant images, 
Places-365~\cite{zhou2017places} and Google Landmark v2~\cite{8237636} contain scene 
images. This allows us to maximize the coverage of our model zoo to different 
domains and enables more effective transfer when fine-tuning on different target 
tasks. 

{\bf Model zoo of single-domain experts.} We build a model zoo of a total of $30$ models 
(Resnet-101~\cite{7780459} and Densenet-169~\cite{HuangLW16a}) trained on $8$ 
large image classification datasets (\ie source datasets). Since each model
is trained on a single classification dataset (\ie domain), we refer to these 
models as single-domain experts. This results in a model zoo, $\mathcal{F} = 
\{f^k\}_{k=1}^{30}$, to evaluate our model selection.

On each source dataset of \tabref{tab:source_dataset}, we train Resnet-101 and 
Densenet-169 models for 90 epochs, with the following hyper-parameters: initial 
learning rate of 0.1, with decay by $0.1\times$ every 30 epochs, SGD with 
momentum of $.9$, weight decay of $10^{-4}$ and a batch size 512. We use the training 
script\footnote{\url{https://bit.ly/38NMvyu}}
from PyTorch~\cite{NEURIPS2019_9015} library and ensure that our models are well-trained. 

In \tabref{tab:source_dataset}, we show slightly higher top-1 test accuracy for our models 
trained on Imagenet~\cite{imagenet_cvpr09} when compared to the PyTorch~\cite{NEURIPS2019_9015}
model zoo\footnote{\url{https://bit.ly/35vZpPE}\label{footnote:torchvision}}. Our 
Resnet-101 model trained on Imagenet has $+.17\%$ top-1 test accuracy and our 
Densenet-169 model has $+.4\%$ top-1 test accuracy vs.\ PyTorch. On source datasets other than 
Imagenet, we train our models with ($\checkmark$) and without ($\times$) Imagenet 
pre-training. This allows us to study the effect of pre-training on a larger dataset 
when we fine-tune and perform model selection. Note that our Resnet-101 models with ($\checkmark$)
Imagenet pre-training have higher accuracy compared to that reported 
in the literature for all source datasets, except iNaturalist~\cite{HornASSAPB17} 
by $-1.03\%$. 

\begin{figure*}[!t]
    \vspace{-.5cm}
    \centering
    \subfigure{\includegraphics[width=.3\textwidth,trim=0 5 0 0,clip]{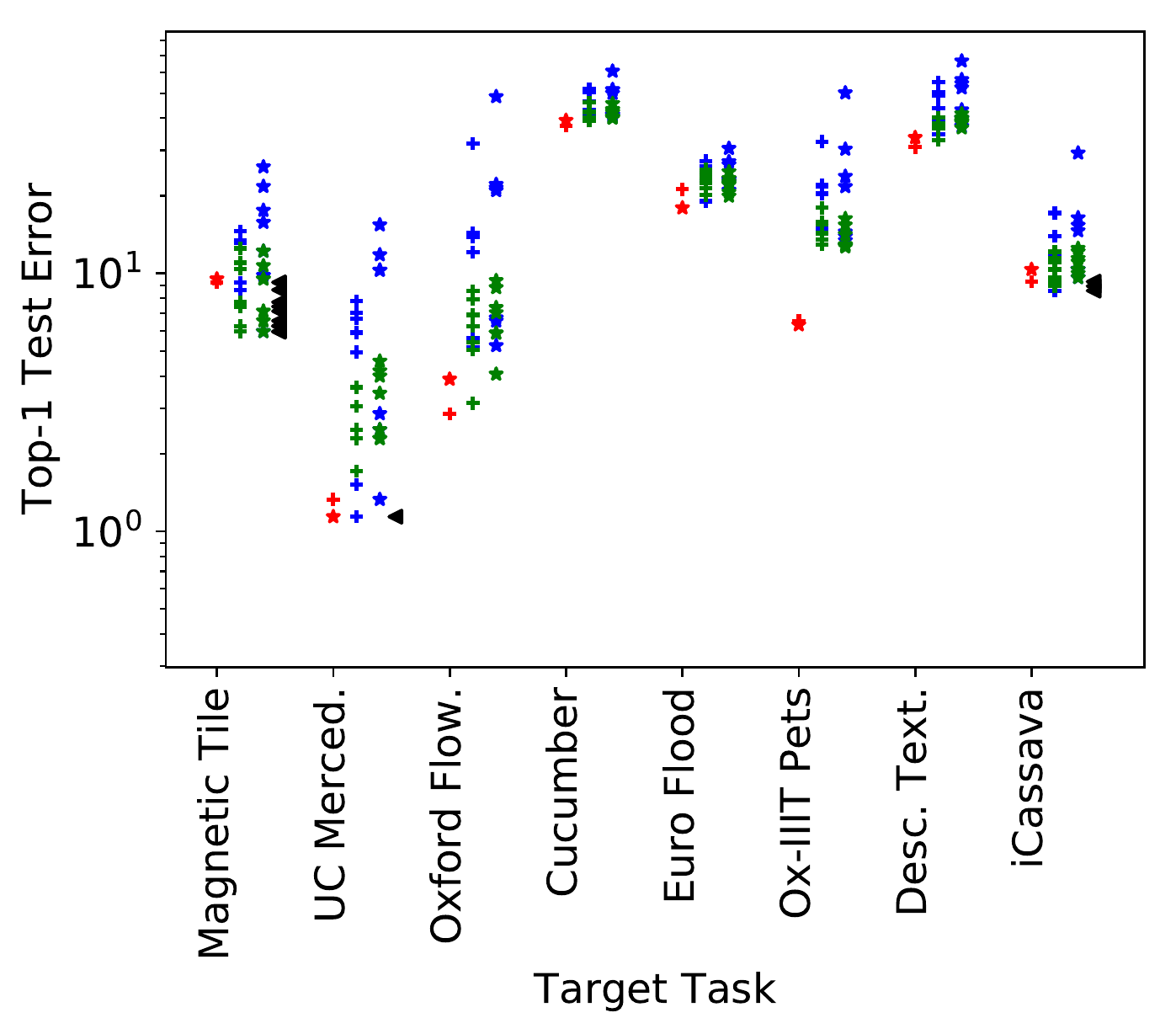}}
    \subfigure{\includegraphics[width=.3\textwidth,trim=0 5 0 0,clip]{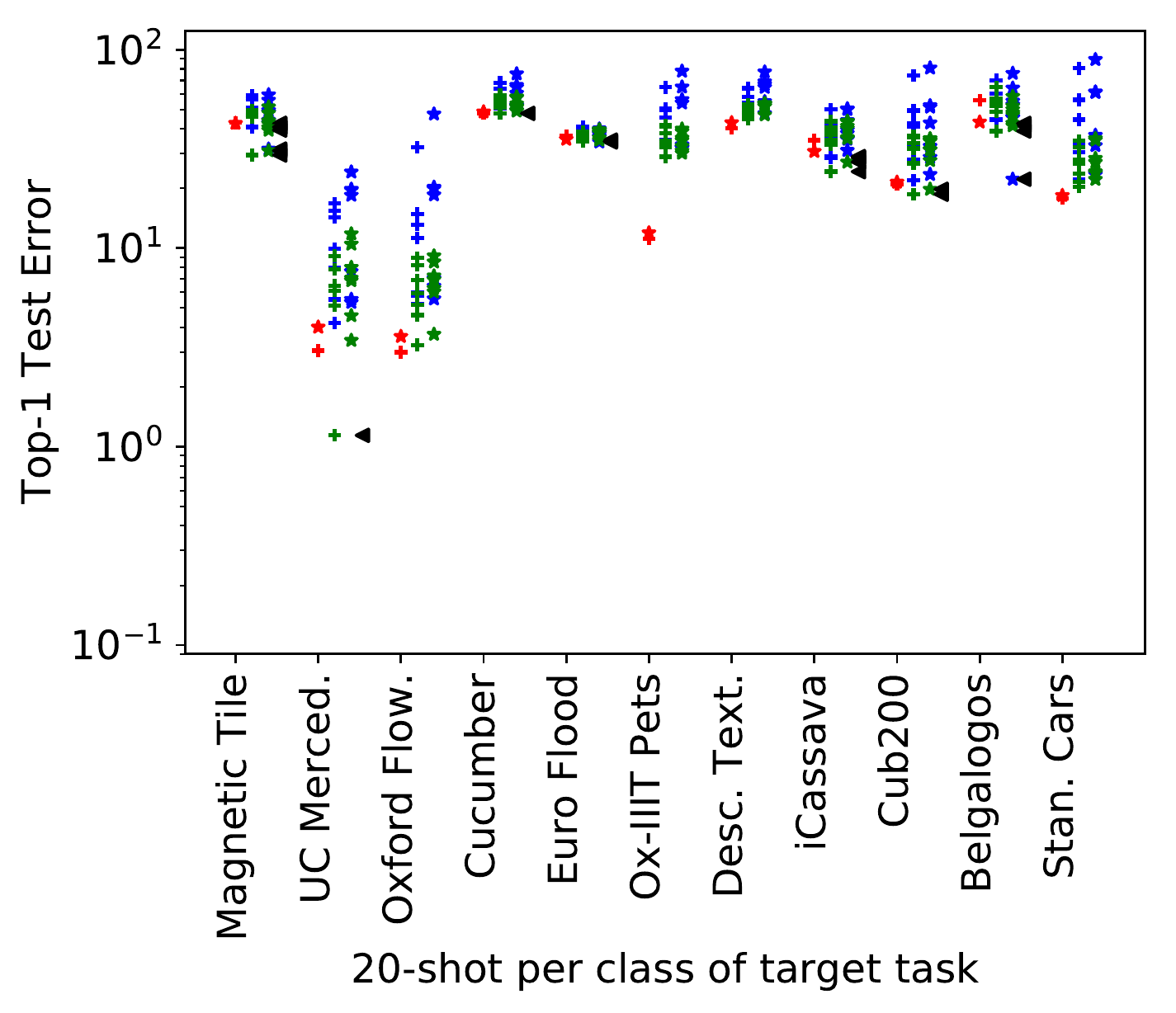}}
    \subfigure{\includegraphics[width=.3\textwidth,trim=0 5 0 0,clip]{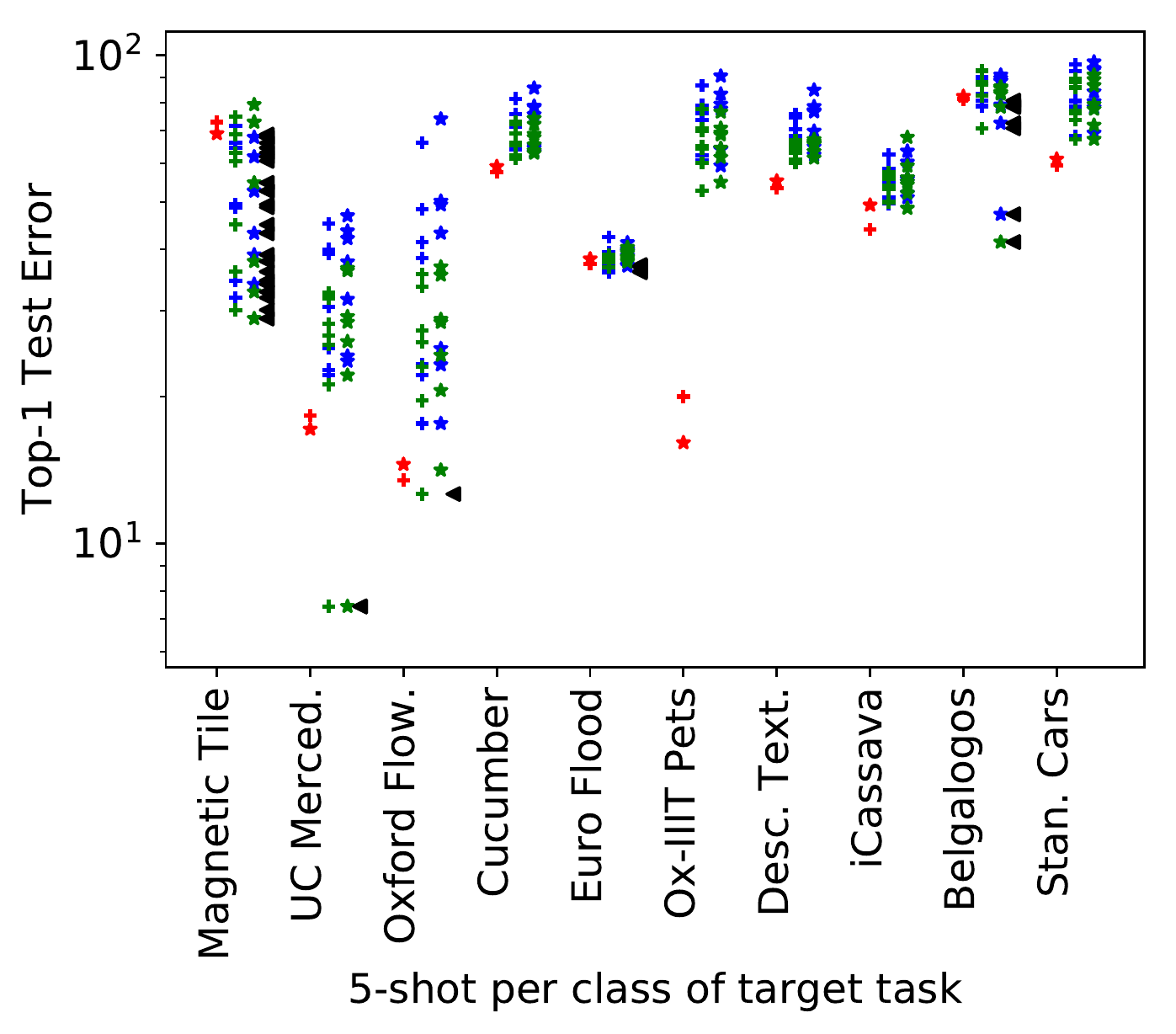}}
    \vspace{-.2cm}
    \subfigure{\includegraphics[width=\textwidth,trim=0 345 0 5,clip]{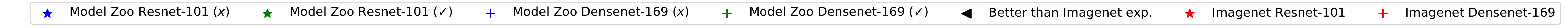}}
\caption{{\bf Fine-tuning with model zoo of single-domain experts.} We plot top-1 test error (vertical axis) for 
fine-tuning with different single domain models in our model zoo. For every target task (on horizontal axis), we have 
$4$ columns of markers from left to right: $1)$ Imagenet experts in \textcolor{red}{red}, $2)$ Densenet-169 experts 
\textcolor{OliveGreen}{with pre-train ($\checkmark$)} and \textcolor{blue}{without pre-train ($\times$)}, $3)$ Resnet-101 experts
\textcolor{OliveGreen}{with pre-train ($\checkmark$)} and \textcolor{blue}{without pre-train ($\times$)}, $4)$ We use ``{\bf black $\leftarrow$}'' 
to highlight models that perform better than imagenet expert (\ie lower error than first column of \textcolor{red}{Imagenet expert} per task). 
Our observations are the following: $i)$ For full target task, we observe better accuracy than Imagenet expert for Magnetic Tile Defects, 
UC Merced Land Use and iCassava (see {\bf black $\leftarrow$}). For 20 and 5-shot per class sampling of target task, with the model zoo we outperform Imagenet expert 
on more datasets, see Oxford Flowers 102, European Flood Depth, Belga Logos and Cub200. Our empirical result, on the importance of 
different pre-trainings of our model zoo experts when training data is limited, adds to the growing body of similar results in 
existing literature~\cite{he2019rethinking,li2020rethinking,zoph2020rethinking}, and
$ii)$ The accuracy gain over Imagenet expert is only obtained for fine-tuning with select few models for a given 
target task, \eg only one expert for UC Merced Land Use target task in Full, 20-shot setting above. Therefore, brute-force fine-tuning with model zoo leads to wasteful computation. Model selection (\secref{sec:approach}) picks the best models to fine-tune and avoids brute-force fine-tuning. Figure is best viewed in high-resolution.
}
\vspace{-.5cm}
\label{fig:finetune_full}
\end{figure*}

\begin{figure}[!t]
\centering
\includegraphics[width=.35\textwidth,trim=0 21 0 0,clip]{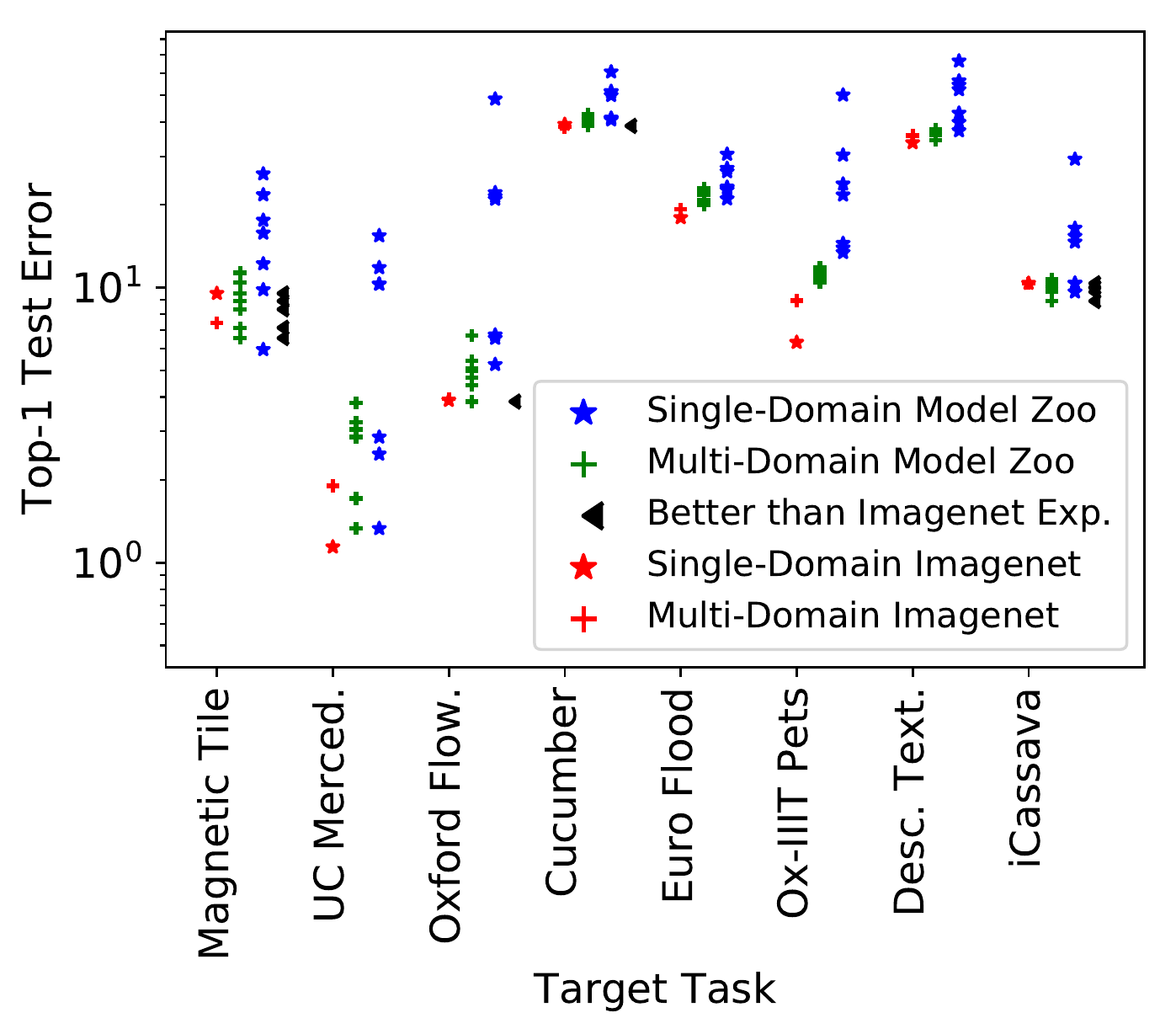}
\caption{{\bf Fine-tuning with the multi-domain expert for the full target task.} 
We use the same notation as \figref{fig:finetune_full}. For every target task (horizontal axis), 
we have $4$ columns corresponding to fine-tuning different models from left to right: $1)$ Imagenet single and 
multi-domain expert in \textcolor{red}{red}, $2)$ Fine-tuning with different domains of multi-domain 
expert in \textcolor{OliveGreen}{green} and $3)$ Single-domain Resnet-101 experts in \textcolor{blue}{blue},
$4)$ We highlight \textcolor{OliveGreen}{multi-domain experts} that obtain lower error than \textcolor{red}{Imagenet} single domain
with {\bf black $\leftarrow$}. Note, since our multi-domain expert is Resnet-101 based, we only 
use all our Resnet-101 experts for for fair comparison. Our observations are: $i)$ We see gains over Imagenet expert 
(both single and multi-domain) by fine-tuning some (not all) domains of the multi-domain expert, for Magentic Tile Defects, 
Oxford Flowers 102, Cucumber and iCassava target tasks. Therefore, it is important to pick the correct domain 
from the multi-domain expert for fine-tuning. $ii)$ We observe the variance in error is smaller for 
fine-tuning with different domains of multi-domain experts, possibly due to shared parameters across 
domains, $iii)$ Finally in some cases, \eg Oxford Flowers 102 and iCassava, our multi-domain experts 
outperform both, all single domain and Imagenet experts. Figure is best viewed in high-resolution.}
\label{fig:finetune_universal}
\end{figure}

{\bf Model zoo of multi-domain expert.} We also train a Resnet-101 based multi-dataset 
(or multi-domain)~\cite{rebuffi-cvpr2018} model on the combination of all the 
$8$ source datasets. Our multi-domain Resnet-101 expert, $f_{w_s, \{w_d\}_{d=1}^{D}}$, 
uses shared weights (or layers), \ie $w_s$, across different domains (or datasets), 
and in addition it has some domain-specific parameters, \ie $\{w_{d}\}_{d=1}^{D}$, 
for each domain. We have 8 source datasets or domains, so $D = 8$ in our benchmark. 
Note, for fine-tuning we can choose any one of the $D$ 
domain-specific parameters to fine-tune. For a given multi-domain expert, 
this results in a model zoo of $D$ models (one per domain) that we can 
fine-tune, $\mathcal{F} = \{f_{w_s, w_1}, f_{w_s, w_2}, \cdots, f_{w_s, w_D}\}$.

We experiment with a few different variants of domain-specific parameters 
-- $i)$ {\bf Shared}: The domain-specific parameters are also shared, 
therefore we simply train a Resnet-101 on all datasets, $ii)$ {\bf Multi-BN}: 
We replace each batch norm in Resnet-101 architecture with a domain-specific 
batch norm. Note, for a batch norm layer we replace running means, scale 
and bias parameters, $iii)$ {\bf Adapter:} We use the domain-specific parallel 
residual adapters~\cite{rebuffi-cvpr2018} within the Resnet-101 architecture. 
Our training hyper-parameters for the multi-domain expert are the same as 
our single-domain expert. The only change is that for every epoch 
we sample at most $100K$ training images (with replacement if $100K$ exceeds
dataset size) from each dataset to balance training between different datasets 
and to keep the training time tractable. As we show in \tabref{tab:multi_domain}, 
{\bf Multi-BN} model outperforms other multi-domain models and we use it in our 
subsequent fine-tuning (\secref{sec:finetune}) and model selection 
(\secref{sec:model_selection}) experiments.

\begin{figure}[!t]
\centering
\subfigure[Full Dataset]{
\includegraphics[width=.228\textwidth]{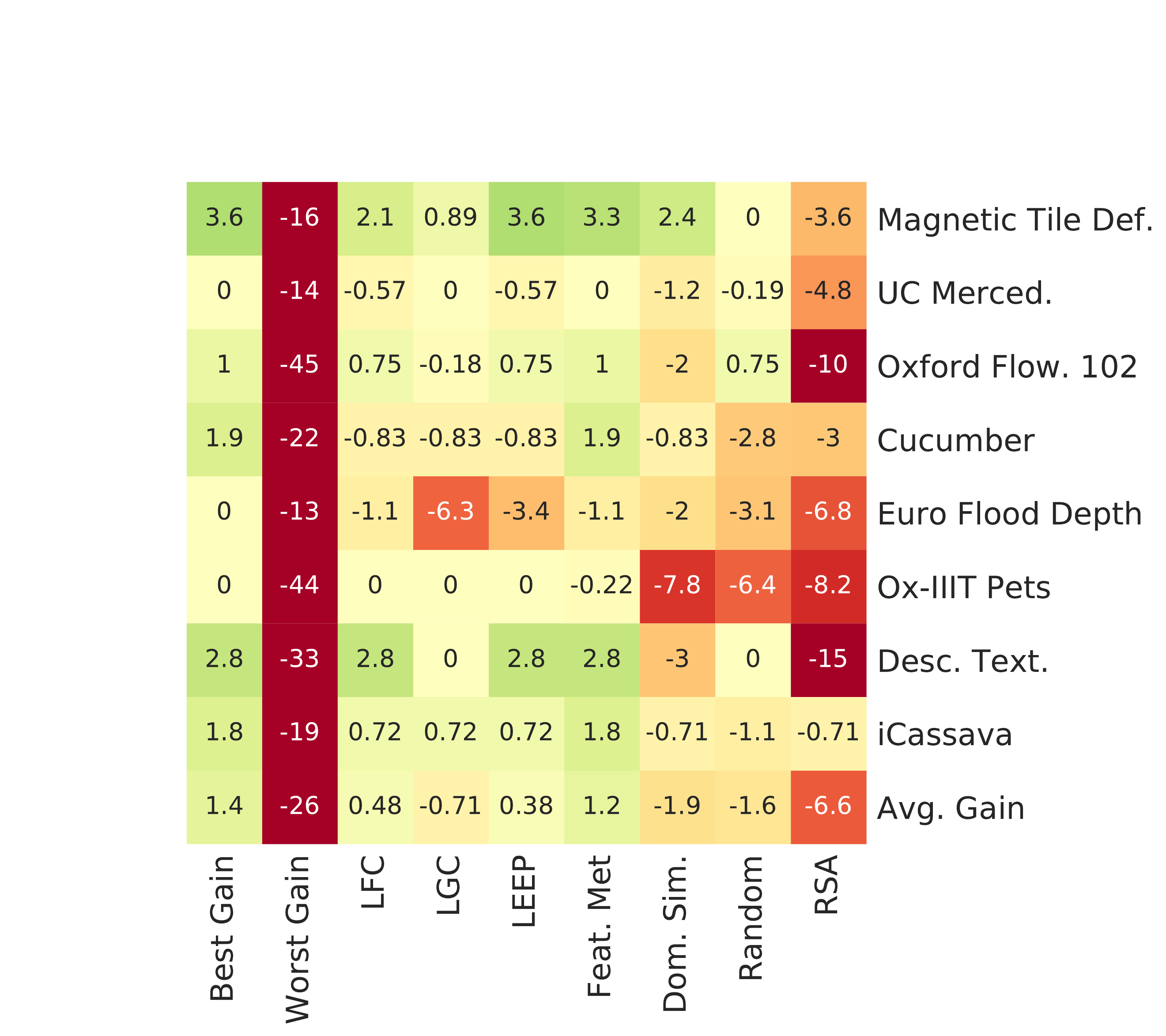}}
\subfigure[20-Shot per class]{
\includegraphics[width=.228\textwidth]{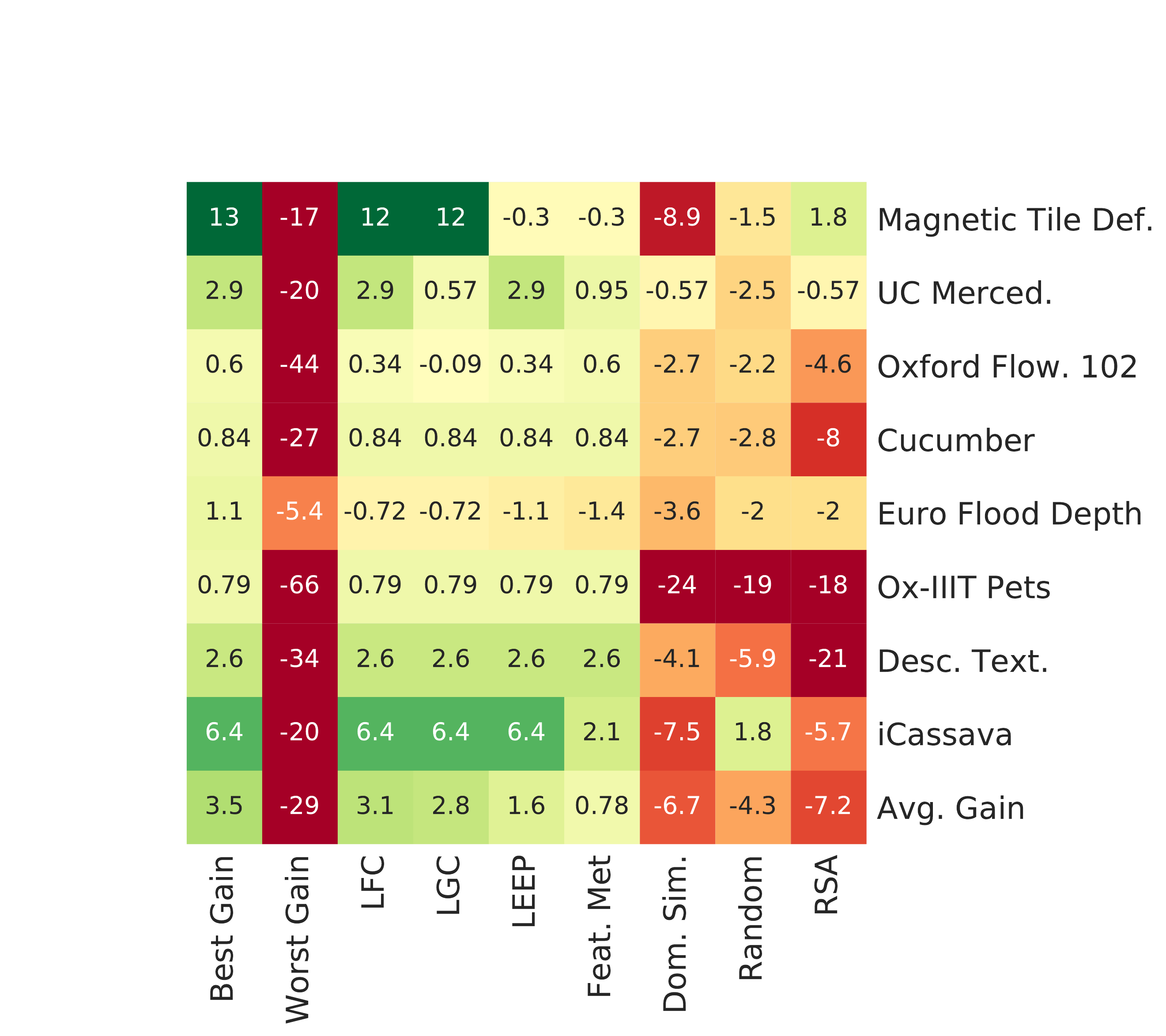}
}
\caption{{\bf Model selection among single-domain experts.} The heatmap shows 
the accuracy gain over Resnet-101 Imagenet expert obtained by fine-tuning the top-$3$ 
selected models for different model selection methods (column) on our target tasks 
(row). \textcolor{OliveGreen}{Higher} values of gain are better. Note, for every method 
we fine-tune all the top-$3$ selected models (with same hyper-parameters as 
\secref{sec:finetune}) and pick the one with the highest accuracy. Model selection performs 
better than ``Worst Gain'' and random selection. On average, LFC, LGC and LEEP~\cite{nguyen2020leep} outperform Domain 
Similarity~\cite{Cui2018iNatTransfer}, RSA~\cite{DwivediR19}. Feature Metrics~\cite{ueno2020a}
performs better than LFC, LEEP in high-data regime, but under-performs in the low-data regime.}
\label{fig:model_zoo_selection_top3}
\end{figure}

\begin{figure}[!t]
    \centering
    \subfigure[Top-1 Selection]{
    \includegraphics[width=.228\textwidth,trim=0 0 0 100,clip]{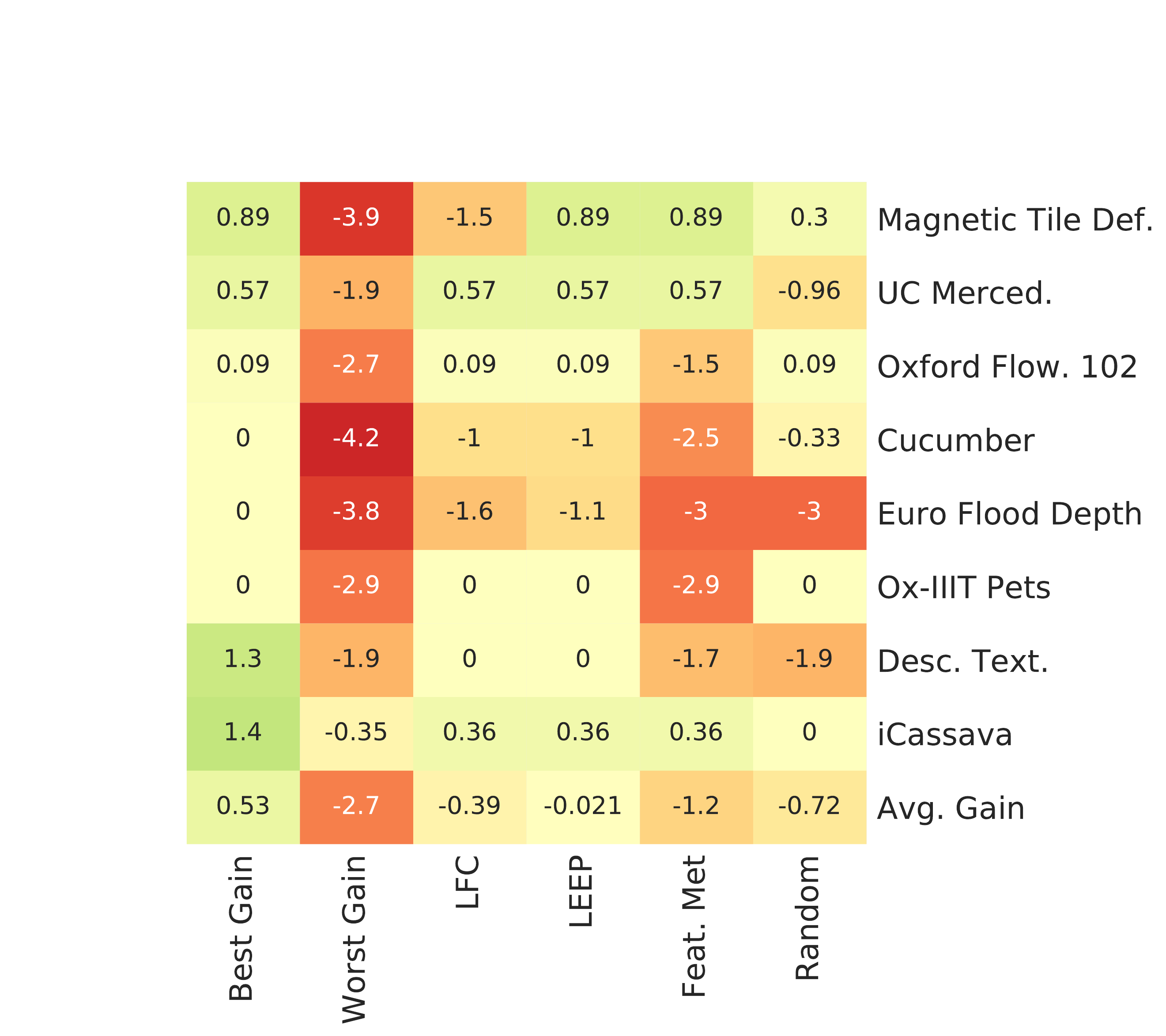}}
    \subfigure[Top-3 Selection]{
    \includegraphics[width=.228\textwidth,trim=0 0 0 100,clip]{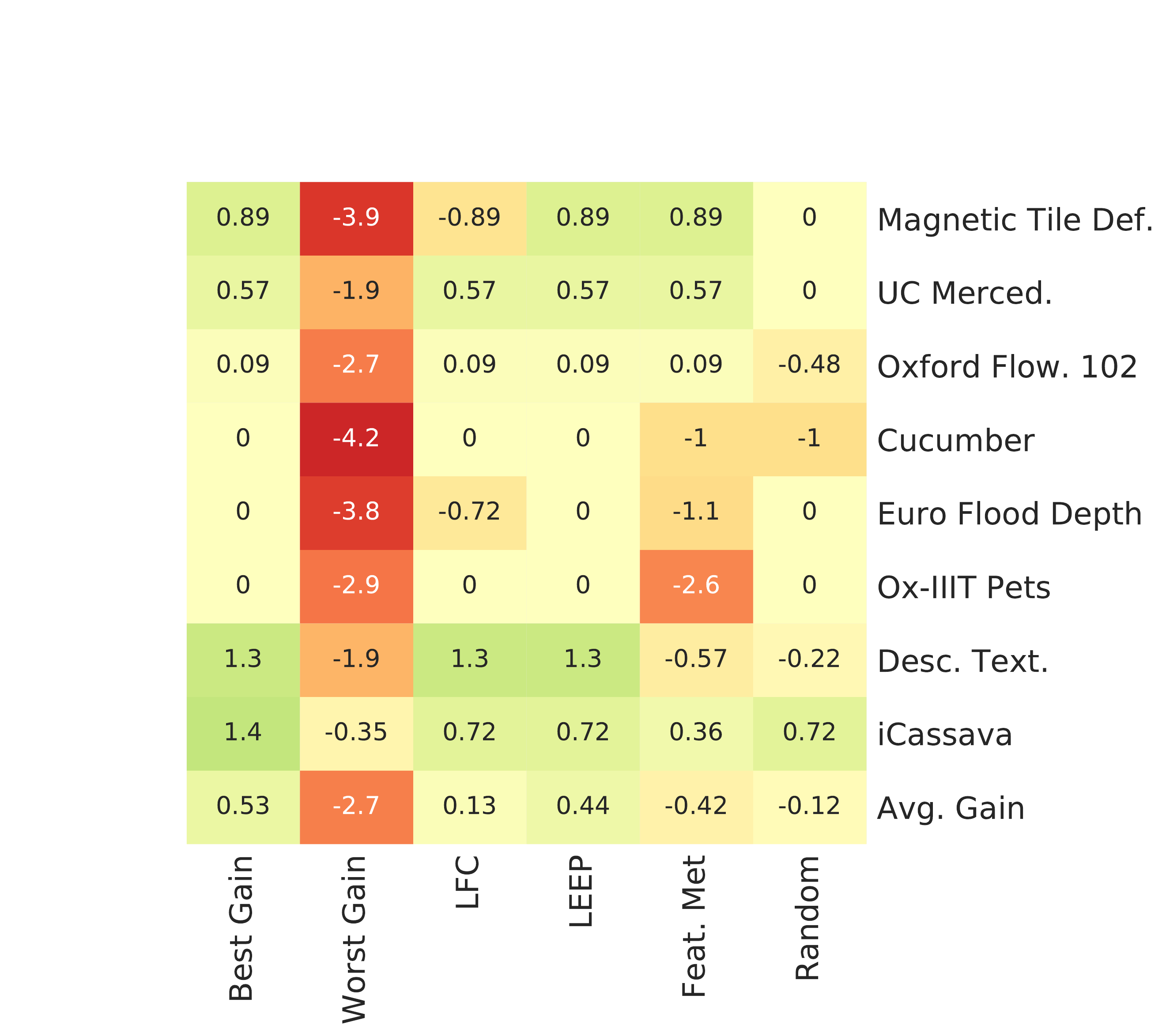}} 
    \caption{{\bf Model Selection with multi-domain expert.} The heatmap shows accuracy
    gain obtained by fine-tuning selected domain over fine-tuning Imagenet domain
    from the multi-domain expert. We show results for top-$1$ and top-$3$ selections.
    LFC, LEEP~\cite{nguyen2020leep} are close to the best gain and they outperform 
    Feature Metrics~\cite{ueno2020a} and \emph{Random}.}
    \label{fig:universal_selection_top1}
\vspace{-.6cm}
\end{figure}

\subsection{Fine-tuning on Target Tasks}
\label{sec:finetune}

{\bf Target Tasks.} We use various target tasks (\tabref{tab:suppl_datasets}) to study transfer 
learning from our model zoo of \secref{sec:model_zoo}: 
Cucumber~\cite{cucumber}, Describable Textures~\cite{cimpoi14describing}, 
Magnetic Tile Defects~\cite{8560423}, iCassava~\cite{mwebaze2019icassava}, 
Oxford Flowers 102~\cite{4756141}, Oxford-IIIT Pets~\cite{parkhi12a}, 
European Flood Depth~\cite{Barz2019EnhancingFI}, UC Merced Land Use~\cite{10.1145/1869790.1869829}. 
For few-shot, due to lesser compute needed, we use additional target
tasks: CUB-200~\cite{WelinderEtal2010}, Stanford Cars~\cite{KrauseStarkDengFei-Fei_3DRR2013}
and Belga Logos~\cite{belgalogos09}. Note, while some target tasks have domain overlap 
with our source datasets, \eg aerial images of UC Merced Land Use~\cite{10.1145/1869790.1869829}, 
other tasks do not have this overlap, \eg defect images in Magnetic Tile Defects~\cite{8560423}, texture images in 
Describable Textures~\cite{cimpoi14describing}.

{\bf Fine-tuning with single-domain experts in model zoo.} For fine-tuning, Imagenet pre-training is a 
standard technique. Note, most deep learning frameworks, \eg PyTorch\ref{footnote:torchvision}, 
MxNet/Gluon\footnote{\url{https://gluon-cv.mxnet.io/api/model_zoo.html}} \etc, just have the Imagenet 
pre-trained models for different architectures in their model zoo. \figref{fig:finetune_full} shows 
the top-1 test error obtained by fine-tuning single-domain experts in our model zoo vs.\ Imagenet 
expert. 

Our fine-tuning hyper-parameters are: $30$ epochs, weight decay of $10^{-4}$, SGD with Nesterov momentum $0.9$, 
batch size of $32$ and learning rate decay by $0.1\times$ at $15$ and $25$ epochs. We observe that 
the most important hyper-parameter for test accuracy is the initial learning rate $\eta$, so for 
each fine-tuning we try $\eta=0.01$,  $0.005$, $0.001$ and report the best top-1 test accuracy. 

\textbf{Does fine-tuning with model zoo perform better than fine-tuning a Imagenet 
expert?} While fine-tuning an Imagenet pre-trained model is standard and works 
well on most target tasks, we show that by fine-tuning models of a large  model-zoo we can indeed 
obtain a lower test error on some target tasks (see models highlighted by {\bf black $\leftarrow$} 
in \figref{fig:finetune_full}). The reduction in error is more pronounced in the low-data 
regime. Therefore, we establish that maintaining a model zoo of models trained on different 
datasets is helpful to transfer to a diverse set of target tasks with different amounts 
of training data.

We demonstrate gains in the low-data regime by training on a smaller subset of the target task, 
with only 20, 5 samples per class in \figref{fig:finetune_full} (\ie, we train in a 20-shot 
and 5-shot setting). In few-shot cases we still test on the full test set.


{\bf Fine-tuning with multi-domain expert.} In~\secref{sec:model_zoo}, we show that
fine-tuning can be done by choosing different domain-specific parameters within 
the multi-domain expert for fine-tuning.  In \figref{fig:finetune_universal}, we fine-tune  
the multi-domain expert, \ie Multi-BN of \tabref{tab:multi_domain}, on our target tasks 
by choosing different domain-specific parameters to fine-tune. Similar to \figref{fig:finetune_full}, we show the accuracy gain obtained by 
fine-tuning multi-domain expert with respect to fine-tuning the standard Resnet-101 pre-trained 
on Imagenet. We observe that selecting the correct domain to fine-tune, \ie the correct 
$w_d$, where $d \in \{1, 2, \cdots, D\}$ from multi-domain model zoo $\mathcal{F} = \{f_{w_s, w_d}\}_{d=1}^{D}$, 
is important to obtain high fine-tuning test accuracy on the target task. In \secref{sec:model_selection},
we show that model selection algorithms help in selecting the optimal 
domain-specific parameters for fine-tuning our multi-domain model zoo. 


We also observe that fine-tuning with our multi-domain expert 
improves over the fine-tuning of single-domain model zoo for
some tasks, \eg iCassava: $+1.4\%$ accuracy gain with multi-domain expert 
compared to $+.72\%$ accuracy gain with single domain model 
expert over Imagenet expert. However, the comparison between 
single domain and multi-domain experts and their transfer properties 
is not the focus of our research and we refer the reader 
to~\cite{Packnet,Rebuffi17,rebuffi-cvpr2018}. 



\begin{table*}[!t]
    \begin{center}
\resizebox{\textwidth}{!}{
    \begin{tabular}{|l||c||ccccc||ccccc|}
    \hline
         {\bf Shots} & {\bf Brute-force} &  \multicolumn{5}{c||}{{\bf Fine-tuning top-3 models}} & \multicolumn{5}{c|}{{\bf Model selection from single-domain model zoo}} \\
         & & {\bf LFC} & {\bf LGC} & {\bf LEEP} & {\bf Feat. Met.} & {\bf Dom.\ Sim.} & {\bf LFC} & {\bf LGC} & {\bf LEEP} & {\bf Feat. Met.} & {\bf Dom.\ Sim.} \\ 
    \hline
    \hline
         Full & 48.17$\times$ & 5.15$\times$ & 3.89$\times$ & 5.01$\times$ & 6.02$\times$ & 4.87$\times$ & .41$\times$ & 8.65$\times$ & .02$\times$ & .00$\times$ & .40$\times$ \\
         20-shot & 41.67$\times$ & 4.35$\times$ & 3.40$\times$ & 3.85$\times$ & 4.86$\times$ & 4.11$\times$ & 1.09$\times$ & 15.26$\times$ & 0.03$\times$ & 0.00$\times$ & 1.31$\times$ \\
    \hline
    \end{tabular}}
    \caption{{\bf Computation cost of model selection and fine-tuning the selected models from single-domain model zoo.} 
    We measure the average run-time for all our target tasks (of \figref{fig:finetune_full}) of: Brute-force fine-tuning
    and Fine-tuning with 3 models chosen by model selection (\figref{fig:model_zoo_selection_top3}). We divide the run-time by the run-time
    of fine-tuning a Resnet-101 Imagenet expert. For the single domain model zoo, brute-force fine-tuning 
    of all $30$ experts requires $40\times$ more computation than fine-tuning Imagenet Resnet-101 expert.
    Note, Densenet-169 models in our model zoo need more computation to fine-tune than Resnet-101, therefore the gain is $>30\times$
    for our model zoo size of $30$.
    With model selection, we can fine-tune with
    selected models in only $3-6\times$ the computation. LFC and LEEP compute model selection scores for $30$ models in
    our zoo with $< 1\times$ the computation of fine-tuning Imagenet Resnet-101 expert. LGC model selection is expensive due to 
    backward passes and large dimension of the gradient vector. However, our LFC approximation to LGC is good at selecting
    models (\figref{fig:model_zoo_selection_top3}) and fast. 
    }
    \label{tab:runtime}
    \end{center}
\vspace{-.6cm}
\end{table*}
\subsection{Model Selection}
\label{sec:model_selection}

\begin{figure}[!t]
    \centering
    \subfigure[Selections for best model]{\includegraphics[width=.23\textwidth]{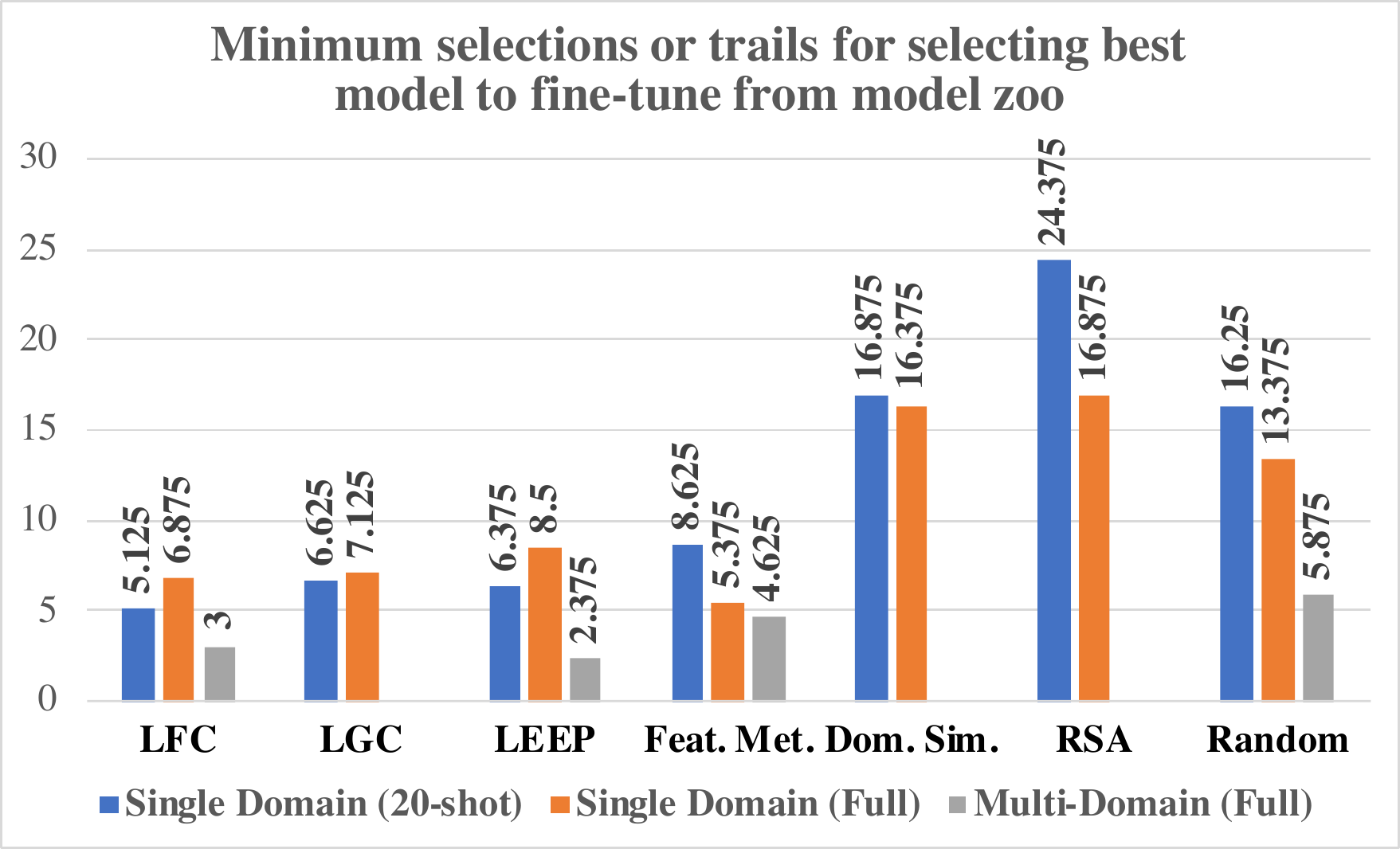}}
    \subfigure[Spearman correlation of expert ranking using model selection scores to actual ranking using fine-tuning accuracy]{\includegraphics[width=.23\textwidth]{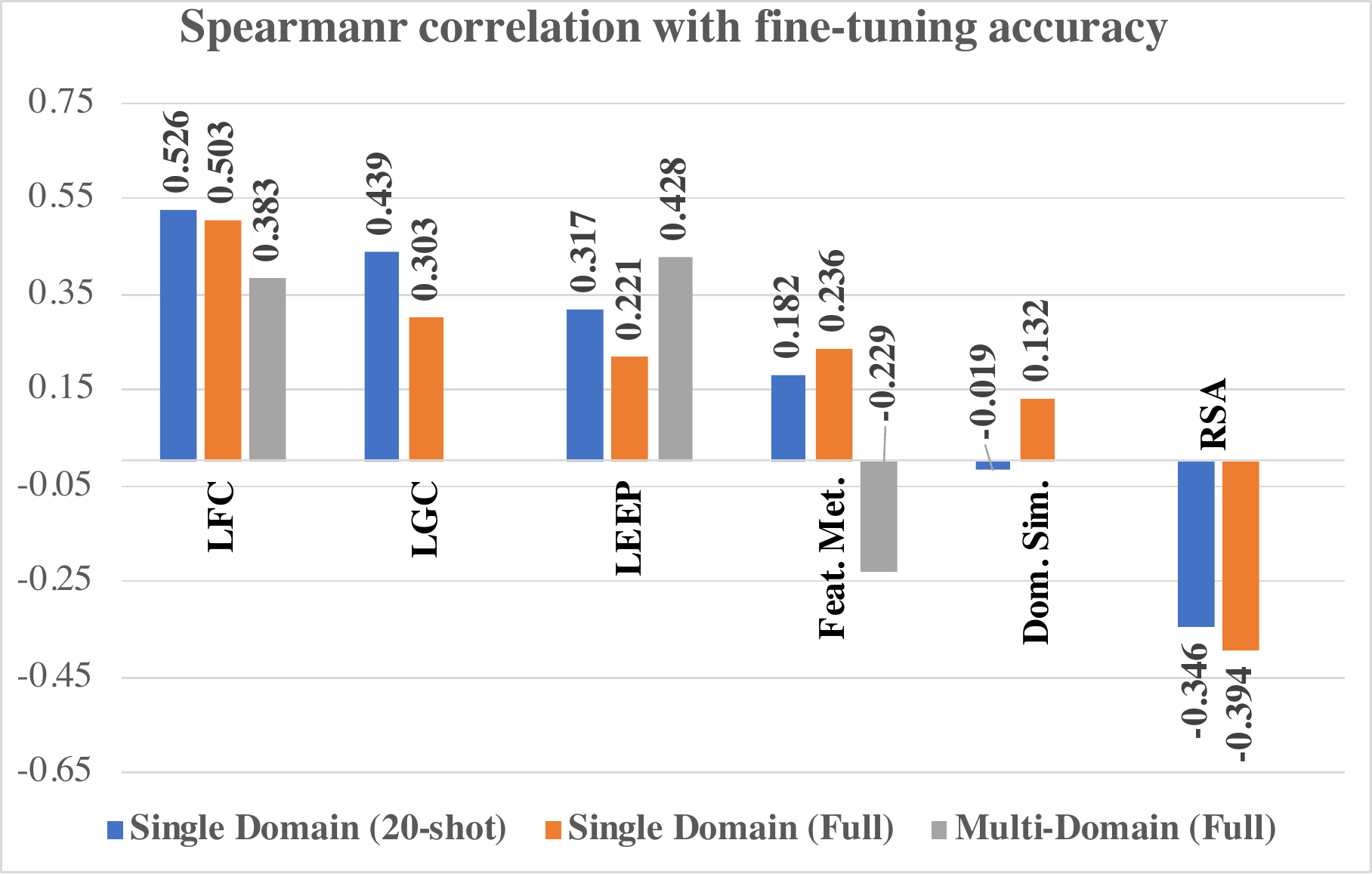}}
    \caption{In $(a)$, we measure the number of trials to select the best model, \ie highest accuracy, from the model zoo. 
    LFC, LGC and LEEP~\cite{nguyen2020leep} require fewer trials than Domain 
    Similarity~\cite{Cui2018iNatTransfer}, RSA~\cite{DwivediR19} and \emph{Random} selection baselines. In $(b)$,
    we show that model selection scores of LFC obtain the highest Spearman's ranking correlation to the actual 
    fine-tuning accuracy compared to other model selection methods. Model selection scores are proxy for 
    fine-tuning accuracy, therefore high correlation is desirable.}
    \label{fig:trials}
\vspace{-.6cm}
\end{figure}

In \secref{sec:finetune}, using our benchmark we find that 
fine-tuning with a model zoo, both single-domain and multi-domain 
domain, improves the test accuracy on the target tasks. Now, we demonstrate 
that using a model selection algorithm we can select the best 
model or domain-specific parameters from our model zoos with only a few 
selections or trials. 


{\bf Model Selection Algorithms.} We use the following scores, $S$, 
for our model selection methods: LFC (see $S_{\text{LF}}$ defined in 
\secref{sec:implementation}), LGC (see $S_{\text{LG}}$ defined 
in ~\eqref{eq:lgc-pearson}), which we introduce in \secref{sec:label-correlation}.
We compare against alternative measures of model selection and/or task 
similarity proposed in the literature: Domain similarity~\cite{Cui2018iNatTransfer}, 
Feature metrics~\cite{ueno2020a}, LEEP~\cite{nguyen2020leep} and RSA~\cite{DwivediR19}.  
Finally, we compare with a simple baseline: \emph{Random} which selects models 
randomly for fine-tuning. 


{\bf Model selection with single-domain model zoo.} In 
\figref{fig:model_zoo_selection_top3}, we select the top-$3$ experts 
(\ie $3$ highest model selection scores) for each model selection method for fine-tuning. We 
do this for all the target tasks (row) using each model selection method 
(column). We use the maximum of fine-tuning test accuracy obtained by 
$3$ selected models to compute accuracy gain with respect to fine-tuning 
with Resnet-101 Imagenet expert. Ideally, we want the accuracy gain with the 
model selection method to be high and equal to the ``Best Gain'' possible for 
the target task. As seen in~\figref{fig:model_zoo_selection_top3}: LFC, LGC and 
LEEP obtain high accuracy gain with just 3 selections in both full dataset 
and 20-shot per class setting. They outperform random selection.

{\bf Model selection with multi-domain expert.} For our multi-domain expert (~\secref{sec:model_zoo}), 
we use model selection to select the 
domain-specific parameters to fine-tune for every model selection method. 
We compute the accuracy gain for fine-tuning using selected domains vs.\ fine-tuning 
Imagenet parameters in the multi-domain expert. It is desirable to have high or close 
to best gain with model selection. Our results in~\figref{fig:universal_selection_top1}, 
show that LFC and LEEP~\cite{nguyen2020leep} obtain higher accuracy gain compared 
to Feature Metrics~\cite{ueno2020a} and \emph{Random} selection.

{\bf Is fine-tuning with model selection faster than brute-force fine-tuning?} 
In~\tabref{tab:runtime}, we show that brute-force fine-tuning is expensive. We can 
save computation by performing model selection using LFC and LEEP and 
fine-tuning only the selected top-$3$ models.

{\bf How many trials to select the model with best fine-tuning accuracy?} In~\figref{fig:trials}, we measure
the average of selections or trials, across all target tasks, required to select the best model 
for fine-tuning from the model zoo. The best model corresponds to the highest fine-tuning 
test accuracy on target task. Our label correlation and LEEP~\cite{nguyen2020leep} methods 
can select the best model in $<7$ trials for our single domain model zoo of $30$ experts 
and in $<3$ trials for the multi-domain model zoo with $8$ domain experts.

{\bf Are model selection scores a good proxy for fine-tuning accuracy?} In~\figref{fig:trials}, we
show our LFC scores have the highest Spearman's ranking correlation to the actual fine-tuning
accuracy for different experts. Note, we average the correlation for all our target tasks. 
Our LFC score is a good proxy for ranking by fine-tuning accuracy and it can
allow us to select (or reject) models for fine-tuning.

\section{Conclusions}

Fine-tuning using model zoo is a simple method to boost accuracy. We 
show that while a model zoo may have modest gains in the high-data 
regime, it outperforms Imagenet experts networks in the low-data regime.
We show that simple baseline methods derived from a linear approximation 
of fine-tuning -- Label-Gradient Correlation (LGC) and Label-Feature Correlation (LFC) -- 
can select good models (single-domain) or parameters (multi-domain) to fine-tune, 
and match or outperform relevant model selection methods in the literature.
Our model selection saves the cost of brute-force fine-tuning and 
makes model zoos viable.






{\small
\bibliography{cvpr}
\bibliographystyle{plain}}

\begin{appendices} 

\begin{table*}[!t]
\resizebox{\textwidth}{!}{
\begin{tabular}{|l||ccc||l|}
\hline
     {\bf Dataset} & {\bf Training Images} & {\bf Testing Images} & {\bf \# Classes} & {\bf URL} \\
\hline
\hline
     NWPU-RESISC45~\cite{7891544} & 25,200 & 6300 & 45 & \footnotesize{\url{https://www.tensorflow.org/datasets/catalog/resisc45}} \\
     Food-101~\cite{bossard14} & 75,750 & 25,250 & 101 & 
     \footnotesize{\url{https://www.tensorflow.org/datasets/catalog/food101}} \\
     Logo 2k~\cite{643999} & 134,907 & 32,233 & 2341 & 
     \footnotesize{\url{https://github.com/msn199959/Logo-2k-plus-Dataset}} \\
     Goog. Landmark~\cite{8237636} & 200,000 & 15,601 & 256 & 
     \footnotesize{\url{https://github.com/cvdfoundation/google-landmark}} \\
     iNaturalist~\cite{HornASSAPB17} & 265,213 & 3030 & 1010 & 
     \footnotesize{\url{https://github.com/visipedia/inat_comp}} \\
     iMaterialist~\cite{imat} & 965,782 & 9639 & 2019 & 
     \footnotesize{\url{https://github.com/malongtech/imaterialist-product-2019}} \\
     Imagenet~\cite{imagenet_cvpr09} & 1,281,167 & 50,000 & 1000 & 
     \footnotesize{\url{http://image-net.org/download}} \\
     Places-365~\cite{zhou2017places} & 1,803,460 & 36,500 & 365 & 
     \footnotesize{\url{http://places2.csail.mit.edu/download.html}} \\
\hline
\hline
    Magnetic Tile Defects~\cite{8560423} & 1008 & 336 & 6 & \footnotesize{\url{https://github.com/abin24/Magnetic-tile-defect-datasets}}\\
    UC Merced Land Use~\cite{10.1145/1869790.1869829} & 1575 & 525 & 21 & \footnotesize{\url{http://weegee.vision.ucmerced.edu/datasets/landuse.html}}\\
    Oxford Flowers 102~\cite{4756141} & 2040 & 6149 & 102 & \footnotesize{\url{https://www.robots.ox.ac.uk/~vgg/data/flowers/102/}} \\
    Cucumber~\cite{cucumber} & 2326 & 597 & 30 & \footnotesize{\url{https://github.com/workpiles/CUCUMBER-9}} \\ 
    European Flood Depth~\cite{Barz2019EnhancingFI} & 3153 & 557 & 2 & \footnotesize{\url{https://github.com/cvjena/eu-flood-dataset}}\\ 
    Oxford-IIIT Pets~\cite{parkhi12a} & 3680 & 3669 & 37 & \footnotesize{\url{https://www.robots.ox.ac.uk/~vgg/data/pets/}}\\
    Describable Textures~\cite{cimpoi14describing} & 4230 & 1410 & 47 & \footnotesize{\url{https://www.robots.ox.ac.uk/~vgg/data/dtd/}}\\
    iCassava~\cite{mwebaze2019icassava} & 5367 & 280 & 5 & \footnotesize{\url{https://sites.google.com/view/fgvc6/competitions/icassava-2019}} \\
    CUB-200~\cite{WelinderEtal2010} & 5994 & 5793 & 200 & \footnotesize{\url{http://www.vision.caltech.edu/visipedia/CUB-200-2011.html}} \\ 
    Belga Logos~\cite{belgalogos09} & 7500 & 2500 & 27 & \footnotesize{\url{http://www-sop.inria.fr/members/Alexis.Joly/BelgaLogos/BelgaLogos.html}} \\
    Stanford Cars~\cite{KrauseStarkDengFei-Fei_3DRR2013} & 8144 & 8041 & 196 & \footnotesize{\url{https://ai.stanford.edu/~jkrause/cars/car_dataset.html}}\\
\hline
\end{tabular}}
\caption{The number of training images, testing images and classes as well as the URL to download the dataset are listed above. The top part contains our source datasets used to train the model zoo and 
the bottom part lists our target tasks used for fine-tuning and model selection with our model zoo.}
\label{tab:suppl_datasets}
\end{table*}

\section{Proofs} 
\label{sec:proof}

\textbf{Proof of Proposition 1.} The proof follows easily from \cite{lee2019wide}. We summarize the steps to make the section self contained.  Assuming, as we do, that the network is trained with a gradient flow (which is the continuous limit of gradient descent for small learning rate), then the weights and activations of the linearized model satisfies the differential equation:
\begin{align*}
    \dot{w}_t &= -\eta \nabla_w f_w(\X)^T \nabla_{f_y(\X)} \mathcal{L} \\
    \dot{f}_t(\X) &= -\eta \nabla_w f_w(\X)^T \nabla_{f_y(\X)} \mathcal{L} = -\eta \Theta \nabla_{f_y(\X)} \mathcal{L}
\end{align*}
For the MSE loss $\mathcal{L} := \sum_{i=1}^N (y_i - f_t^\lin(x_i))^2$, the second differential equations become a first order linear differential equation, which we can easily solve in close form. The solution is
\[
f_t^\lin(\mathcal{X}) = (I - e^{-\eta \Theta t}) \Y + e^{-\eta \Theta t} f_0(\X).
\]
Putting this result in the expression for the loss at time $t$ gives
\begin{align*}
\mathcal{L}_t &= \sum_{i=1}^N (y_i - f_t^\lin(x_i))^2\\
&=(\Y - f_t^\lin(\X))^T(\Y - f_t^\lin(\X))     \\
&=(\Y - f_{w_0}(\X))^T e^{-\eta \Theta t}(\Y - f_{w_0}(\X)),
\end{align*}
as we wanted.

\textbf{Proof of using feature approximation in kernel.} Using the notation $\E_{i,j}[a_{ij}] := \frac{1}{N^2} \sum_{i,j=1}^N a_{ij}$ we have
\begin{align*}
\y^T \,  \Theta \, \y &= N^2 \E_{i,j}[ y_i y_j \Theta_{ij}]\\
&=  N^2 \E_{i,j}[ y_i y_j \nabla_w f_w(x_i) \cdot \nabla_w f_w(x_j)]
\end{align*}
Now let's consider an $f_w$ in the form of a DNN, that is $f_w(x) = W_L \phi( W_{L-1} \ldots \phi(W_0 x))$. By the chain rule, the gradient of the weights at layer $l$ is given by:
\[
\nabla_{W_l} f_w(x) = J_{l+1}(x) \otimes f^l_w(x)
\]
where $J_{l+1}$ is the gradient of the output pre-activations coming from the upper layer and  $f^l_w(x)$ are the input activations at layer $l$ and ``$\otimes$'' denotes the Kronecker's  product or, equivalently since both are vectors, the outer product of the two vectors.
Recall that $\|A \otimes B\|_2= \|A\|_2 \|B\|_2$, which will be useful later.
Using this, we can rewrite $\y^T \,  \Theta \, \y $ as follows:
\begin{align*}
&\y^T \,  \Theta \, \y \\
&= N^2\E_{i,j}[ y_i y_j \nabla_w f_w(x_i) \cdot \nabla_w f_w(x_j)] \\
&= N^2\E_{i}[ y_i \nabla_w f_w(x_i)] \cdot \E_j[y_j \nabla_w f_w(x_j)] \\
&=  N^2\sum_{l=1}^L  \E_{i}[ y_i J_{l+1}(x_i) \otimes f^l_w(x_i)] \cdot \E_{j}[ y_j J_{l+1}(x_j) \otimes f^l_w(x_j)]
\end{align*}

We now introduce a further approximation and assume that $J_{l+1}$ is uncorrelated from  $f^l_w(x_i)$. The same assumption is used by \cite{martens2015optimizing} (see Section 3.1) who also provide theoretical and empirical justifications. Using this assumption, we have:
\begin{align*}
\y^T \,  \Theta \, \y &= N^2
\sum_{l=1}^L \Big\| \E_{i}[ y_i J_{l+1}(x_i) \otimes f^l_w(x_i)] ] \Big\|^2 \\
&= N^2 \sum_{l=1}^L \Big\|   \E_{i}[ J_{l+1}(x_i) ] \otimes \E_{i}[ y_j f^l_w(x_j) ] \Big\|^2 \\
&= N^2 \sum_{l=1}^L \Big\|   \E_{i}[ J_{l+1}(x_i) ] \Big\|^2 \Big\|\E_{i}[ y_j f^l_w(x_j) ] \Big\|^2
\end{align*}

The term $\E_{i}[ y_j f^l_w(x_j)]$ measures the correlation between each individual feature and the label. If features are correlated with labels, then $\y^T \,  \Theta \, \y$ is larger, and hence initial convergence is faster.
Note that we need not consider only the last layer, convergence speed is determined by the correlation at all layers. Note however that the contribution of earlier layers is discounted by a factor of $\|\E_{i}[ J_{l+1}(x_i)]\|^2$. As we progress further down the network, the average of the gradients may become increasingly smaller, decreasing the term $\|\E_{i}[ J_{l+1}(x_i)]\|^2$ and hence diminishing the contribution of earlier layer clustering to convergence speed.

\section{Datasets}

We choose our source and target datasets such that they cover different domains, and 
are publicly available for download. Detailed data statistics are in the respective 
citations for the datasets, and we include a few statistics \eg training images, 
testing images, number of classes in \tabref{tab:suppl_datasets}. For all the 
datasets, if available we use the standard train and test split of the dataset,
else we split the dataset randomly into 80\% train and 20\% test images. If
images are indexed by URLs in the dataset, we download all accessible URLs 
with a python script.

\section{Details of model selection methods}
\label{sec:model_selection_impl}
\begin{figure*}
    \subfigure[25 samples per class]{
    \includegraphics[width=.3\textwidth]{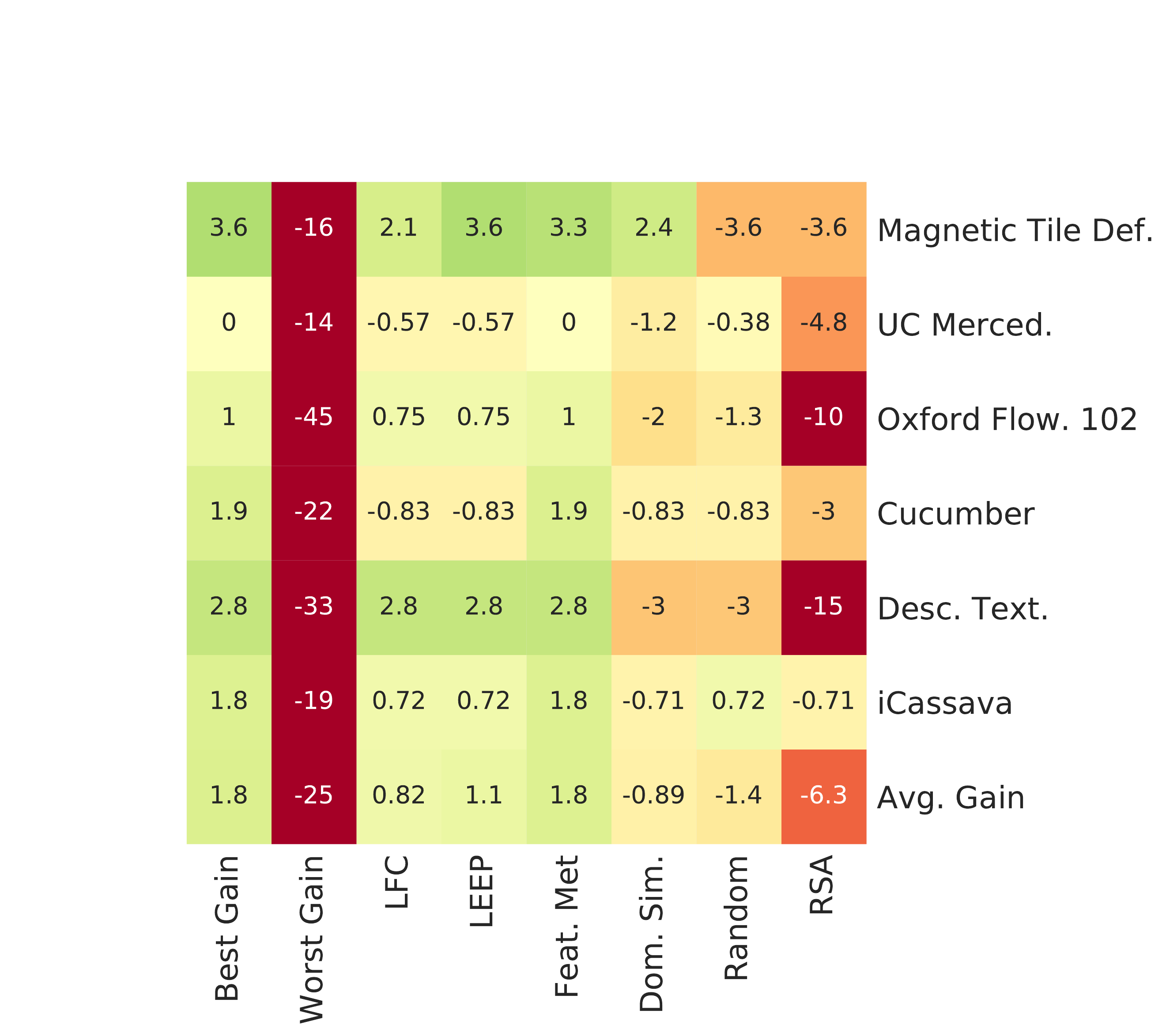}}
    \subfigure[50 samples per class]{
    \includegraphics[width=.3\textwidth]{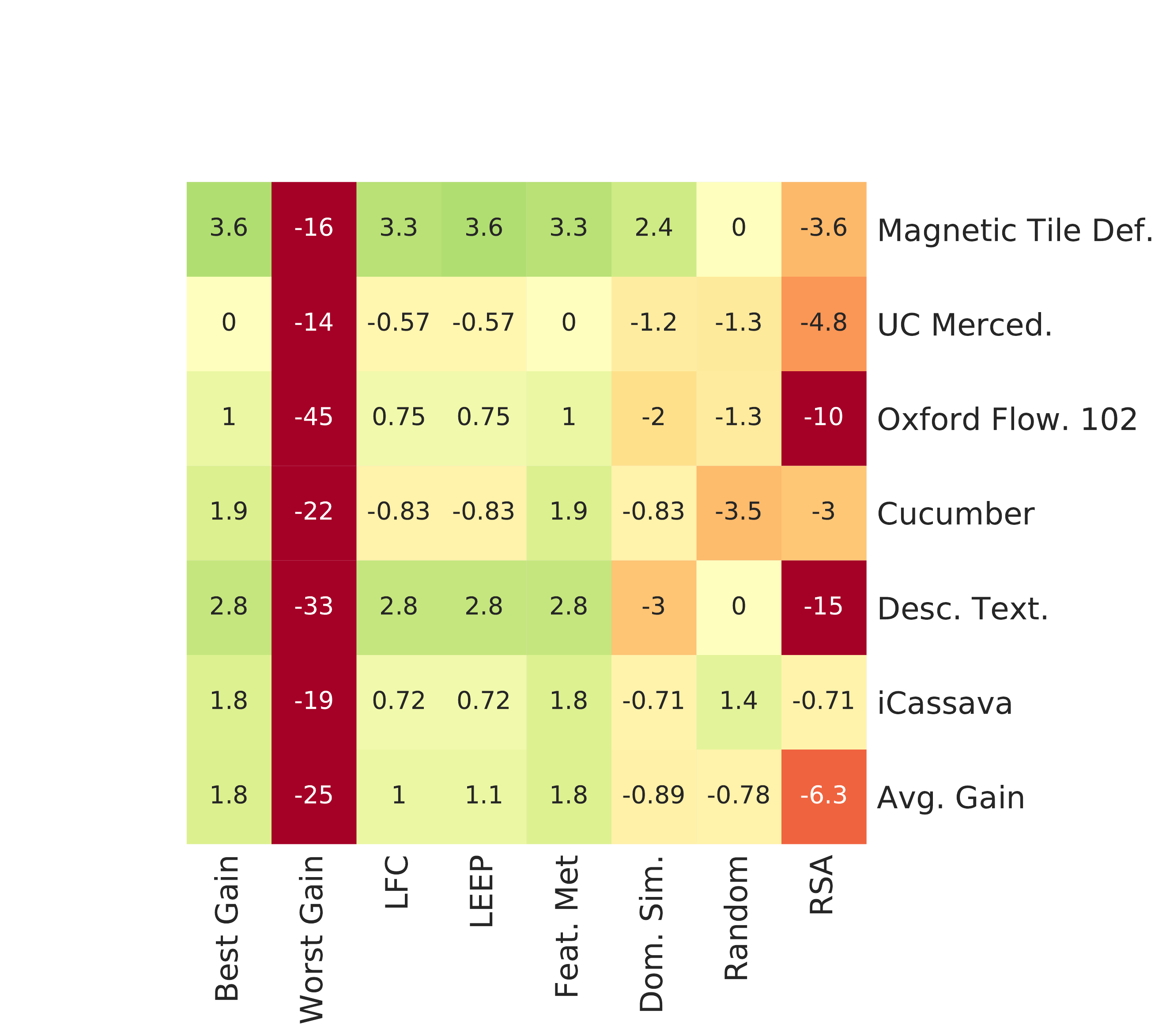}}
    \subfigure[Full target task]{
    \includegraphics[width=.3\textwidth]{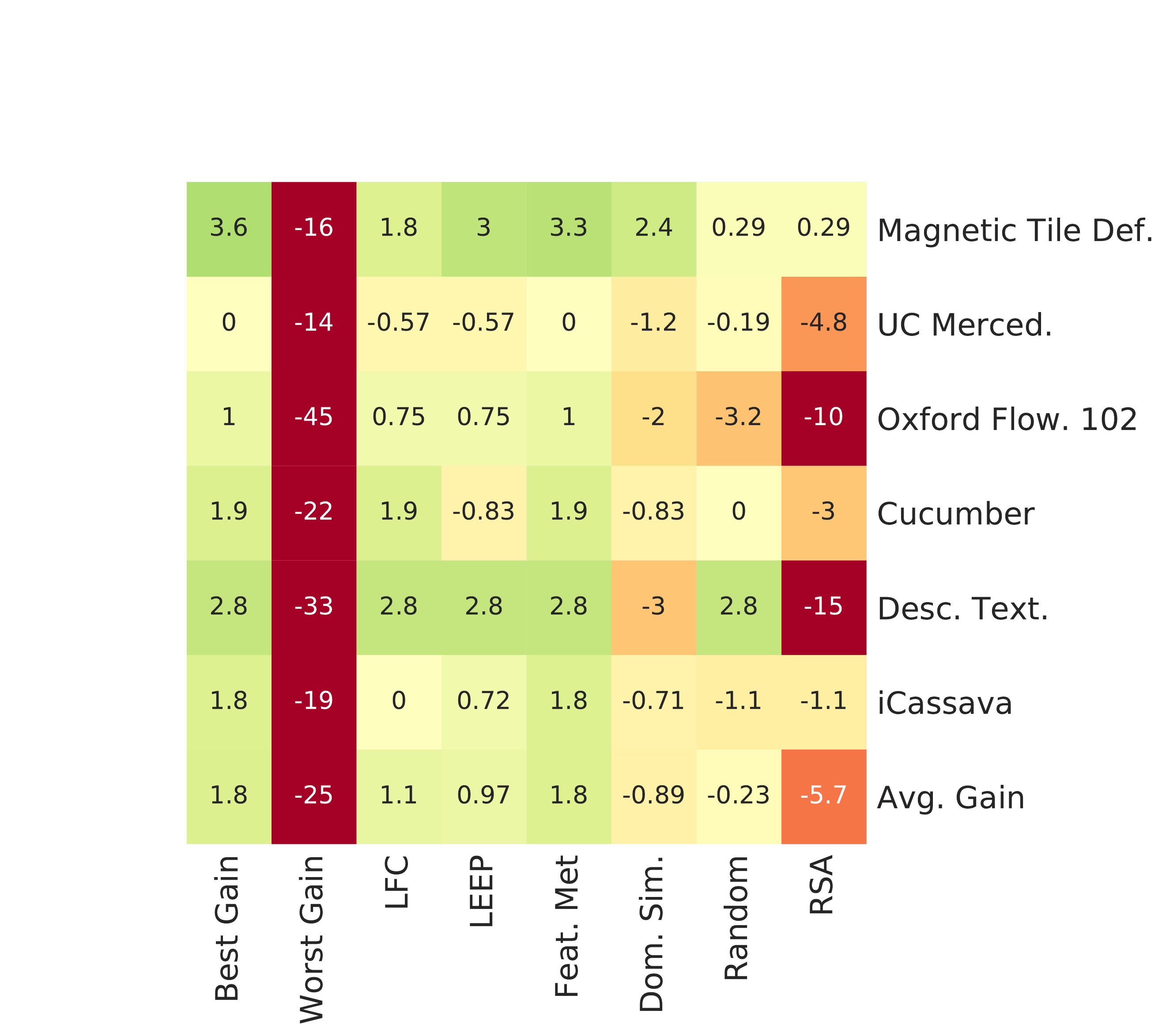}}
    
    \caption{{\bf Ablation study of dataset size for model selection.} Above we use 25,50-samples per class and full target task to perform model selection with different methods. We plot accuracy gain vs. Imagenet expert for top-3 selected models for every method (similar to Fig.\ 4  of the paper). The accuracy gain increases for LFC, LEEP and RSA with more samples of the target task. However, we see that even as small as $25$ samples suffice to obtain good accuracy gain with low computational cost.}
    \label{fig:model_selection_dataset_size}
\end{figure*}

\begin{figure*}
    \subfigure[NWPU-RESISC45, Epoch=0]{
    \includegraphics[width=.3\textwidth]{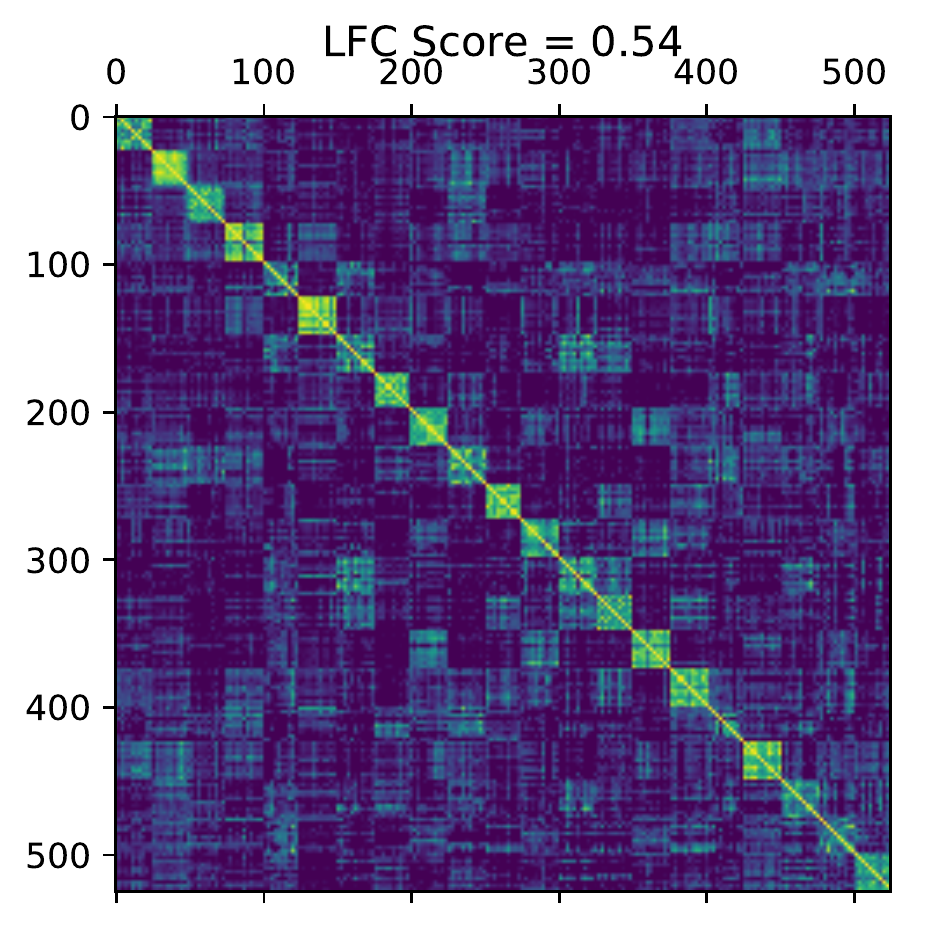}}
    \subfigure[NWPU-RESIC45, Epoch=15]{
    \includegraphics[width=.3\textwidth]{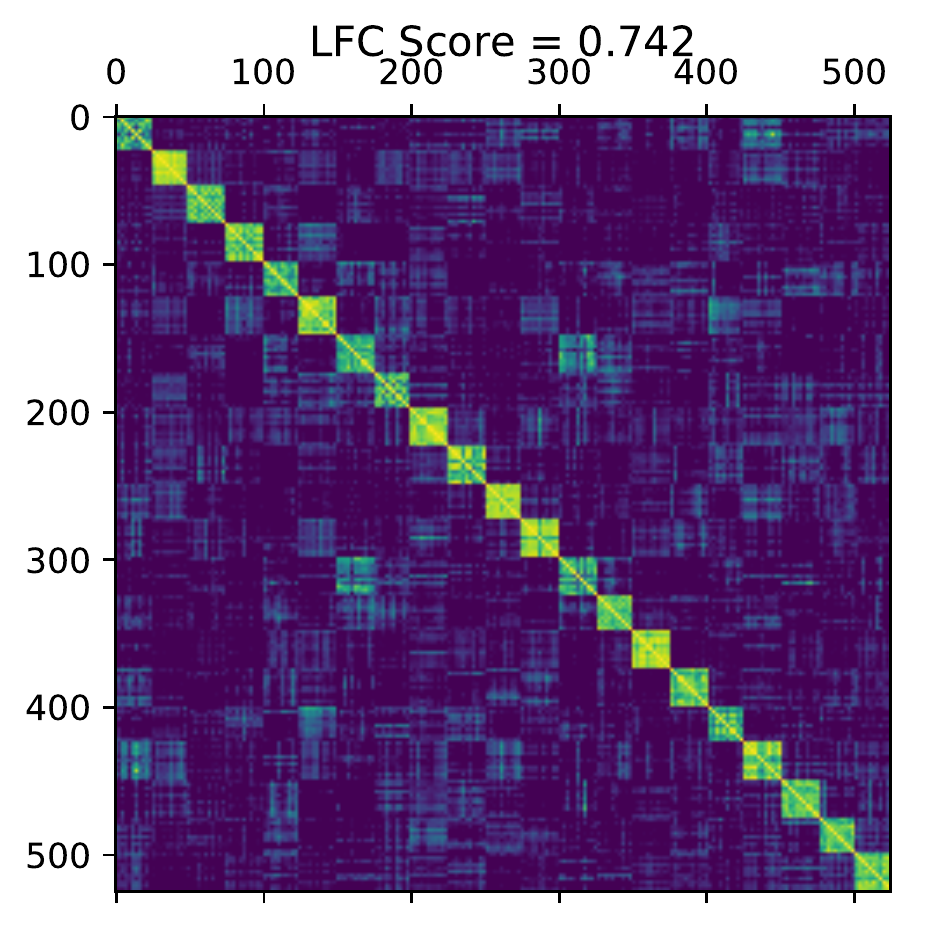}}
    \subfigure[NWPU-RESISC45, Epoch=30]{
    \includegraphics[width=.3\textwidth]{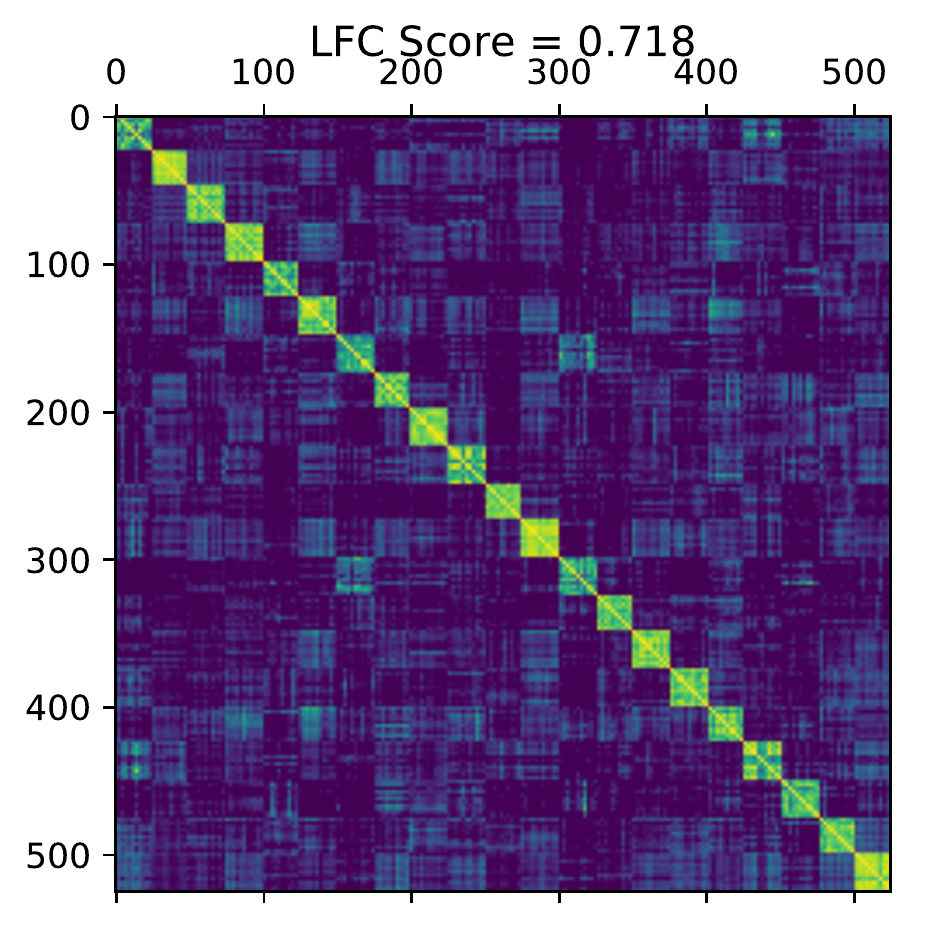}}
    
    \subfigure[Food-101, Epoch=0]{
    \includegraphics[width=.3\textwidth]{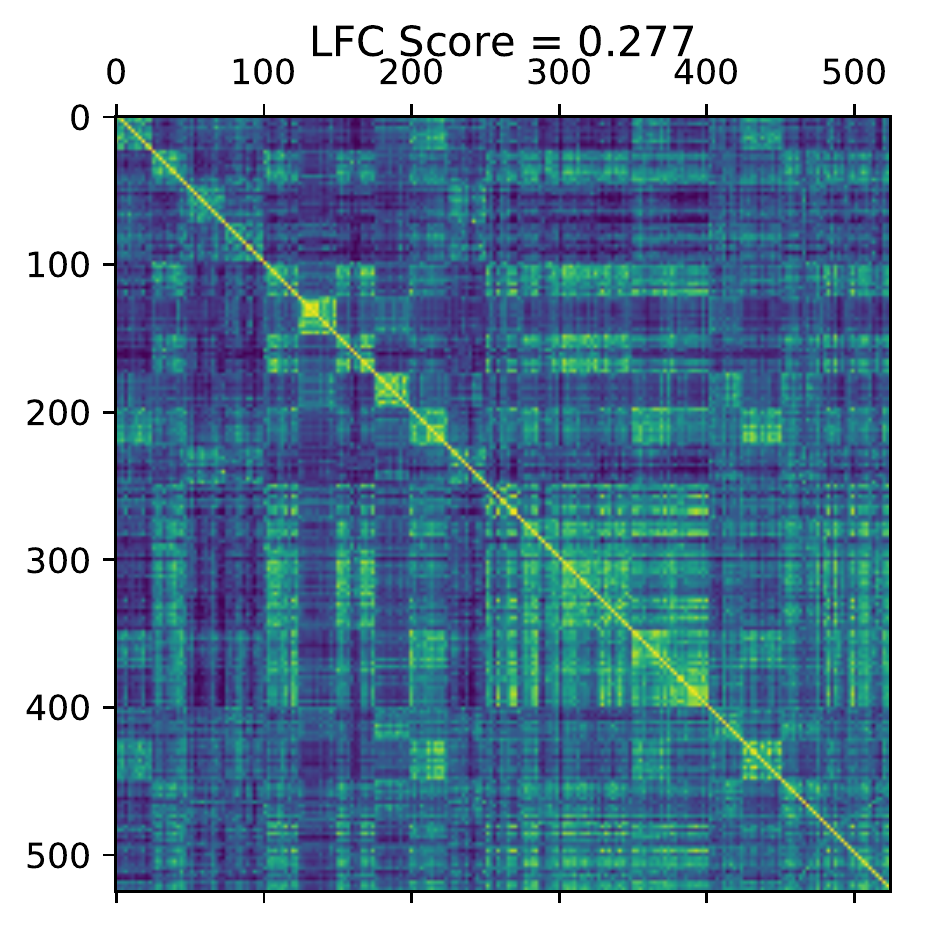}}
    \subfigure[Food-101, Epoch=15]{
    \includegraphics[width=.3\textwidth]{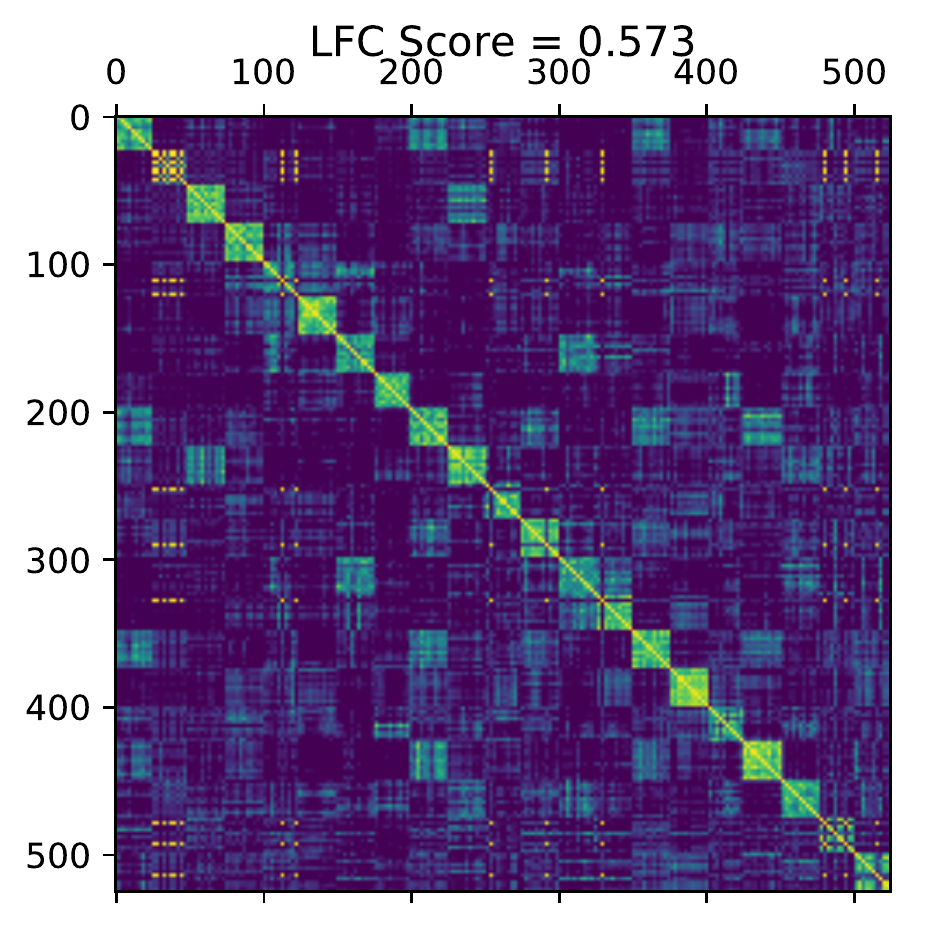}}
    \subfigure[Food-101, Epoch=30]{
    \includegraphics[width=.3\textwidth]{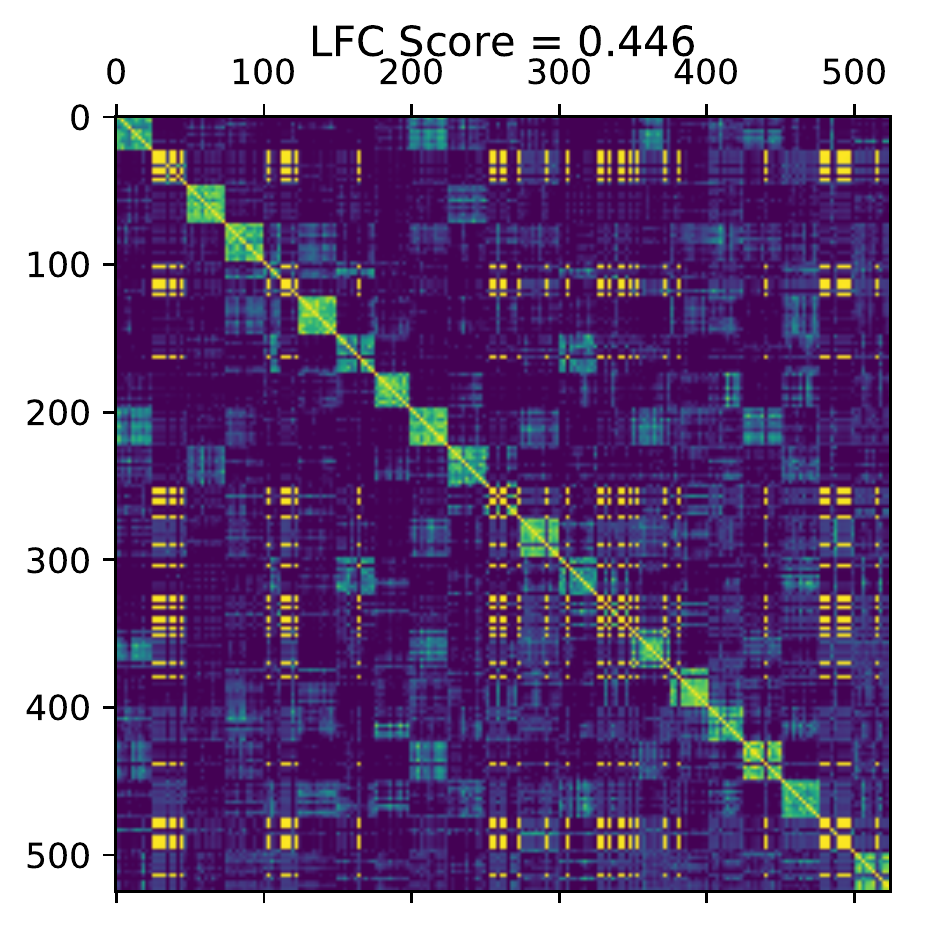}}
    
    \subfigure[Logo 2k, Epoch=0]{
    \includegraphics[width=.3\textwidth]{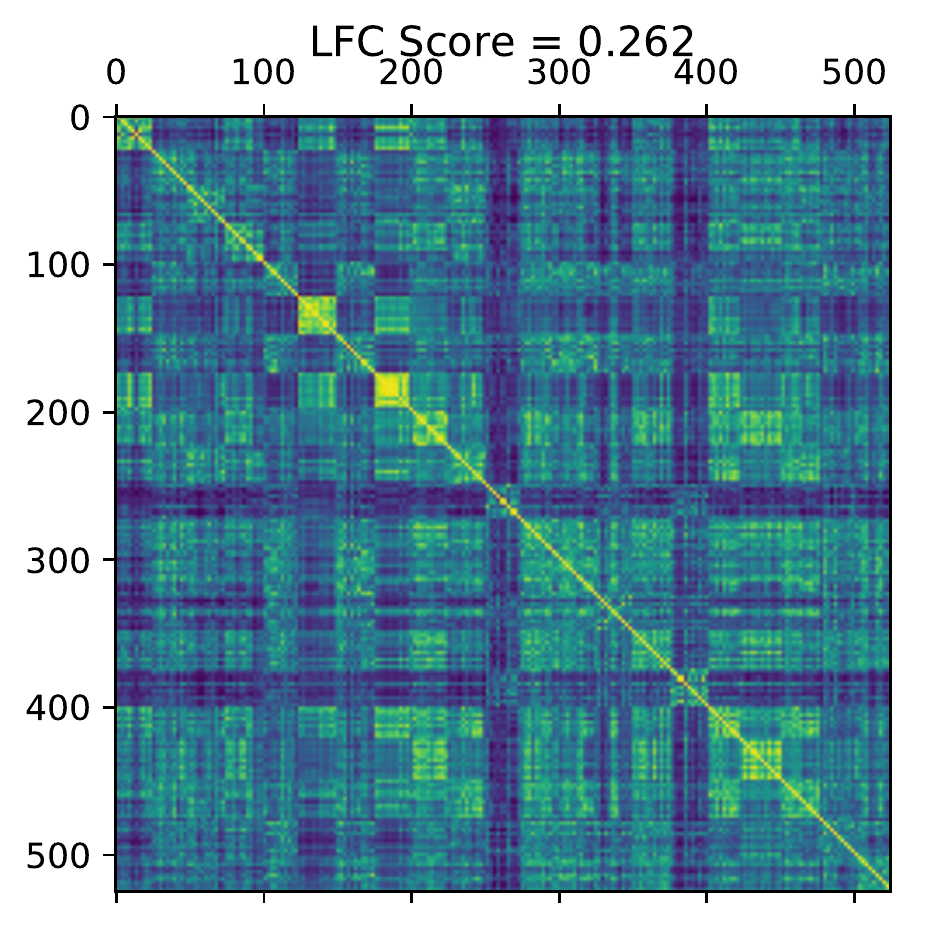}}
    \subfigure[Logo 2k, Epoch=15]{
    \includegraphics[width=.3\textwidth]{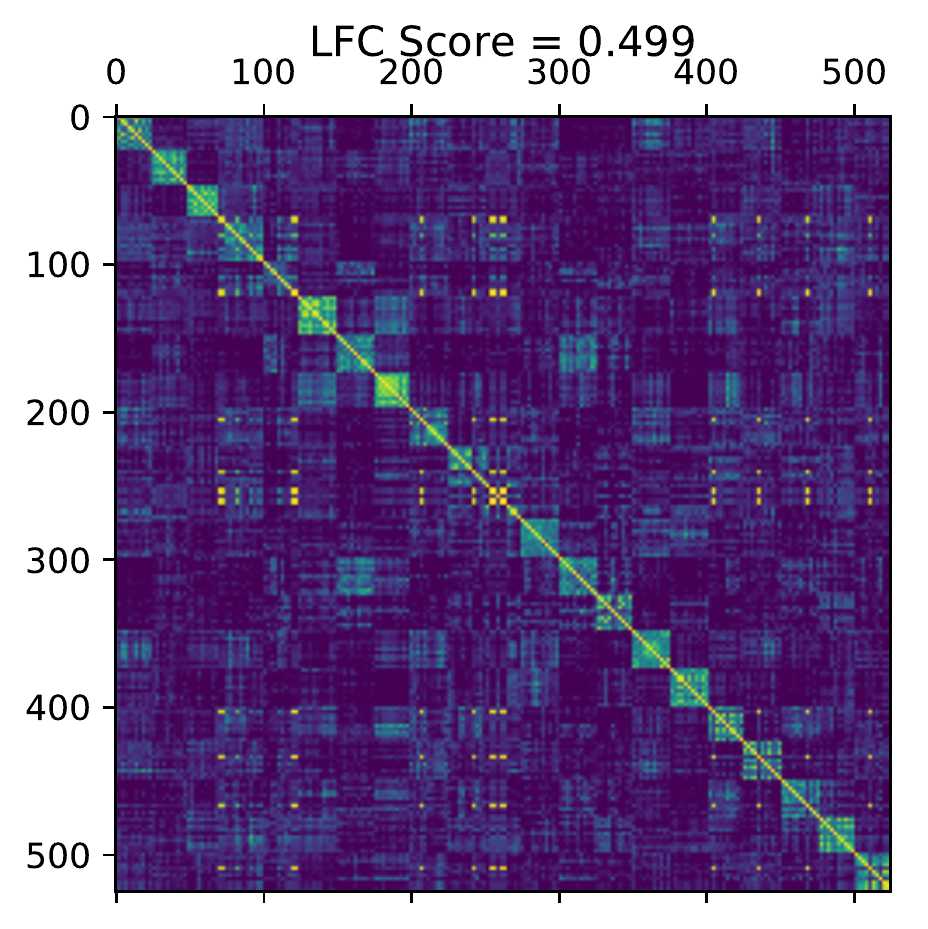}}
    \subfigure[Logo 2k, Epoch=30]{
    \includegraphics[width=.3\textwidth]{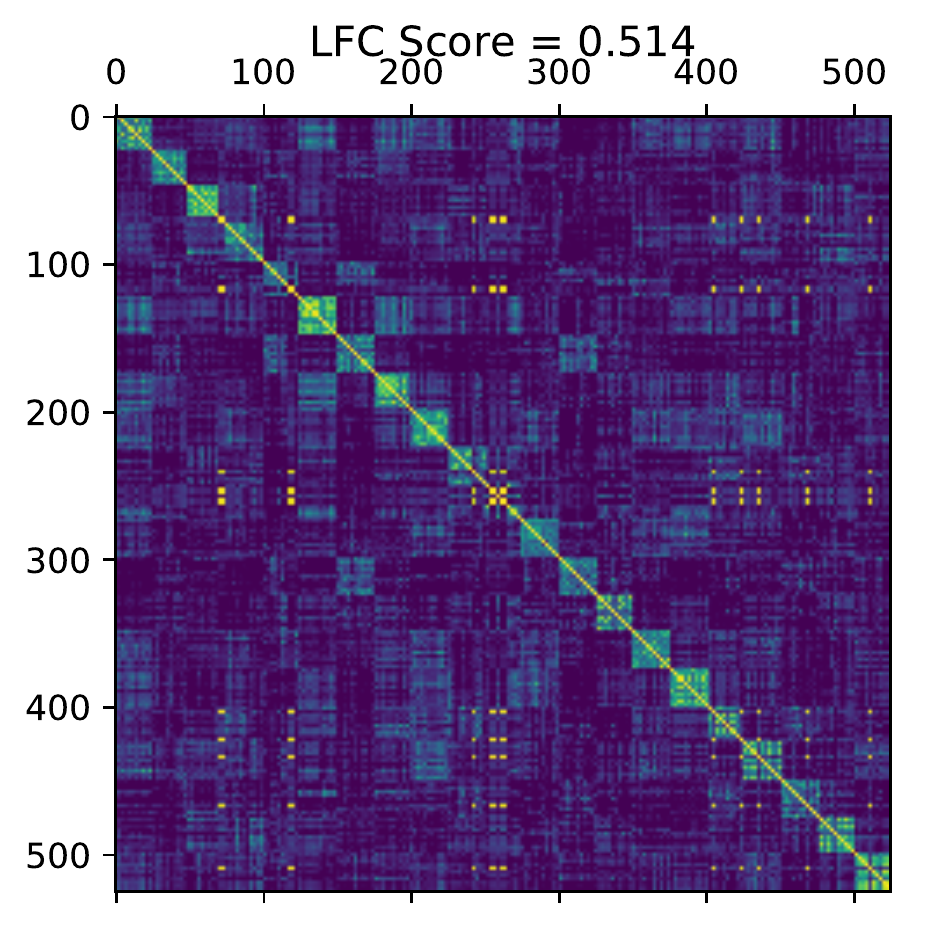}}
\vspace{-.3cm}    
    \caption{We plot the \emph{feature correlation matrix}, $\Theta_F$, for different pre-trainings (row) and different epochs (columns) of fine-tuning. Above, we fine-tune
    on the UC Merced Land Use~\cite{10.1145/1869790.1869829} dataset comprising of aerial images. Images with same class label, 25 images per class, are grouped along the vertical/horizontal axis. Since, features of the same class should be correlated and features of different classes should be uncorrelated, the matrix is expected to have higher values along block diagonal and zero elsewhere. We observe that the matrix exhibits this ideal behaviour
    for pre-training on semantically related domain (aerial images) of NWPU-RESISC45~\cite{7891544} (top row) and has highest LFC score for this pre-training.}
    \label{fig:rdm_visualize}
    \vspace{-.5cm}
\end{figure*} 
{\bf Domain Similarity \cite{Cui2018iNatTransfer}.} As per ~\cite{Cui2018iNatTransfer}, 
we extract avg. features for every class for source and target datasets using 
pre-trained model. We compute an earth movers distance between these average class 
vectors and convert them to domain similarity score. We use the code provided by the authors at
\url{https://github.com/richardaecn/cvpr18-inaturalist-transfer}. We exclude classes with 
less than $5$ training images for Earth-Movers Distance computation.

{\bf RSA \cite{DwivediR19}.} Following the procedure outlined in~\cite{DwivediR19}, 
we extract features before the classification layer 
(\eg 2048 dim features of Resnet-101 after average pool) for images in the target 
dataset. We denote this set of features as $f(x)$, $\forall (x, y) \in \mathcal{D}$.
We build a representation dissimilarity matrix (RDM) as follows: 

\begin{equation}
\text{rdm}_{f}(i, j) = 1 - correlation(f(x_{i}), f(x_{j}))
\end{equation}

We train a small neural network $f_{\text{small}}$ on target dataset.
Note, this is much cheaper to train than fine-tuning the model zoo. 
Features are extracted from $f_{\text{small}}$ and we build another
rdm: 

\begin{equation}
\text{rdm}_{f_{\text{small}}}(i, j) = 1 - correlation(f_{\text{small}}(x_{i}), f_{\text{small}}(x_{j}))
\end{equation}

If rdm's of trained small network $f_{\text{small}}$ and our pre-trained model 
$f$ are similar, then the pre-trained model is a good candidate for fine-tuning 
with target dataset. The final RSA model selection score is:

\begin{equation}
S_{\text{RSA}}(f, \mathcal{D}) = \text{spearmanr}(\text{rdm}_{f}, \text{rdm}_{f_{\text{small}}})
\end{equation}

Since the method requires training a small neural network on target task, we train a Resnet-18 as 
the small neural network with the same fine-tuning configuration used in Section 4.1 of the paper 
with initial learning rate = $.005$.

{\bf Feature Metrics \cite{ueno2020a}.} Features are extracted for all images of
target dataset from pre-trained model, \ie $f(x), \forall x \in \mathcal{D}$. 
We use same features as RSA, our LFC/LGC and compute variance, sparsity metrics of ~\cite{ueno2020a}. 
We use the sparsity metrics as model selection score, $S_{\text{Feat. Metrics}}(f, \mathcal{D}) 
= \text{sparsity}({f(x), \forall x \in \mathcal{D}})$. Note, we use the optimal linear combination 
of the two sparsity metrics proposed in the paper. For feature metrics, the hypothesis is that if 
the pre-trained model generates more sparse representations, they are can generalize with 
fine-tuning to the target task.

{\bf LEEP~\cite{nguyen2020leep}.} LEEP builds an empirical classifier from source
dataset label space to target dataset label space using base model $f$. 
The likelihood of target dataset $\mathcal{D}$ under this empirical classifier is the model 
selection score for the pre-trained model and target dataset. See~\cite{nguyen2020leep} for 
a detailed explanation.

\section{Different dataset size for model selection}

In~\figref{fig:model_selection_dataset_size}, we perform an ablation study on different sampling size of the target task used for model selection. We find that, our choice of $25$ samples per class for model selection, suffices to select good models to fine-tune in top-3 selections at low-computational cost.

\section{Visualization of $\Theta_F$ with fine-tuning} 

In~\figref{fig:rdm_visualize}, we plot the feature correlation
matrix for different pre-trained models across different epochs
of fine-tuning (\ie $0^{th}$, $15^{th}$, $30^{th}$ epoch) for the UC Merced Land Use~\cite{10.1145/1869790.1869829}
target task. We see that the pre-trained model on NWPU-RESISC45~\cite{7891544}, exhibits the ideal correlation wherein 
features of the images with the same class are correlated and features of images with different classes are uncorrelated. This NWPU-RESISC45~\cite{7891544} also
has the highest LFC score.

\end{appendices}

\end{document}